\documentclass{article}

\usepackage[preprint]{neurips_2020}
\usepackage{float}
\usepackage{graphicx,wrapfig}
\usepackage[utf8]{inputenc}
\usepackage[T1]{fontenc}    
\usepackage{hyperref}      
\usepackage{url}          
\usepackage{booktabs}     
\usepackage{amsfonts}       
\usepackage{nicefrac}       
\usepackage{microtype}      
\usepackage{mathtools}
\usepackage{tikz}
\usepackage{graphicx}
\usepackage{pgfplots}
\usepackage{multirow}
\usepackage{subcaption}
\usepgfplotslibrary{groupplots}
\usepackage{natbib}
\usepackage{bm,bbm}
\usepackage[ruled,vlined]{algorithm2e}

\newcommand\R{\mathbb{R}}
\usepackage{tikz}
\usetikzlibrary{positioning}
\usetikzlibrary{shapes.geometric, arrows}
\tikzstyle{Relu} = [isosceles triangle, draw=red,rotate=90, minimum size =0.5cm,fill=blue!20]
\tikzstyle{HRelu} = [isosceles triangle, draw=red,dashed,rotate=90, minimum size =0.5cm,fill=blue!20]
\tikzstyle{linear} = [circle, minimum size =0.5cm, draw=red, fill=red!30]
\tikzstyle{linear1} = [circle, minimum size =0.8cm, draw=red, fill=red!30]
\tikzstyle{box} = [rectangle, minimum width =1cm, minimum height =0.5cm, draw=black,dotted, fill=green!30]
\tikzstyle{box_weight} = [rectangle, minimum width =1cm, minimum height =0.5cm, draw=red, fill=red!30]
\tikzstyle{input} = [rectangle, minimum width =1cm, minimum height =0.5cm, draw=white, fill=white]
\tikzstyle{arrow} = [thick,->,>=stealth]

\tikzstyle{bias} = [circle, draw=red, fill=red!30,minimum size = 0.2cm, inner sep=0pt]

\pgfplotsset{compat=1.16}
\DeclareMathOperator*{\argmax}{arg\,max}
\title{A neural network based model for multi-dimensional nonlinear Hawkes processes }

\author{
	Sobin Joseph\\
	Department of Management Studies\\
	Indian Institute of Science\\
	Bangalore 560012\\
	\texttt{sobinjoseph@iisc.ac.in}\\
	\And
	Shashi Jain\\
	Department of Management Studies\\
	Indian Institute of Science\\
	Bangalore 560012\\
	\texttt{shashijain@iisc.ac.in} \\
}
\begin{document}
	
	\maketitle
	
	\begin{abstract}
		
		This paper introduces the Neural Network for Nonlinear Hawkes processes (NNNH), a non-parametric method based on neural networks to fit nonlinear Hawkes processes. Our method is suitable for analyzing large datasets in which events exhibit both mutually-exciting and inhibitive patterns. The NNNH approach models the individual kernels and the base intensity of the nonlinear Hawkes process using feed forward neural networks and jointly calibrates the parameters of the networks by maximizing the log-likelihood function. We utilize Stochastic Gradient Descent to search for the optimal parameters and propose an unbiased estimator for the gradient, as well as an efficient computation method. We demonstrate the flexibility and accuracy of our method through numerical experiments on both simulated and real-world data, and compare it with state-of-the-art methods. Our results highlight the effectiveness of the NNNH method in accurately capturing the complexities of nonlinear Hawkes processes.
	\end{abstract}


	\section{Introduction}
	\label{Sec:intro}

	A.G. Hawkes introduced the Hawkes process, which is a type of multivariate point process used to model a stochastic intensity vector that depends linearly on past events \cite{Hawkes1971spectra}. This approach has been widely applied in various fields, such as seismology \cite{ogata1999seismicity, marsan2008extending}, financial analysis \cite{filimonov2012quantifying, bacry2015Hawkes}, social interaction modeling \cite{crane2008robust, blundell2012modelling, zhou2013learning}, criminology \cite{mohler2011self}, genome analysis \cite{reynaud2010adaptive, carstensen2010multivariate}, and epidemiology \cite{park2020non, chiang2022Hawkes}.
	
	The Hawkes process in its original form is linear, i.e., the intensities depend on past events through a linear combination of kernel functions.  The primary concern in modelling the Hawkes process is estimating these kernel functions.  A common practice has been to assume a parametric form for the kernel function, such as the exponential and power-law decay kernels, and then use maximum likelihood estimation \citep{ozaki1979maximum} to determine the optimal values of the parameters.  
	
	While easier to estimate, a parametric kernel function is often not expressive enough for real-world problems. A significant body of work is concerned with the non-parametric estimation of the kernel functions. \cite{lewis2011nonparametric} propose a non-parametric, Expectation-Maximization (EM) based method to estimate the Hawkes kernel and the non-constant base intensity function. \cite{zhou2013learning}  uses the method of multipliers and majorization-minimization approach to estimate the multivariate Hawkes kernels. \cite{bacry2014second} develop a non-parametric estimation based on the Wiener-Hopf equations for estimating the Hawkes kernels. \cite{achab2017uncovering} derive an integrated cumulants method for estimating the Hawkes kernel.  \cite{xu2016learning} proposed a sparse-group-lasso-based algorithm combined with likelihood estimation for estimating the Hawkes kernels. \cite{joseph2022shallow} use a feed-forward network to model the excitation kernels and fit the network parameters using a maximum likelihood approach. Except for \cite{bacry2014second}, these non-parametric kernel estimation methods are generally restricted to the linear Hawkes processes.  
	
	Modelling the intensities as a linear combination of kernel functions imposes a non-negativity constraint on the kernel functions, which can be interpreted as excitation kernels. Non-negative kernel functions do not allow incorporating inhibitive effects while modelling the intensity process. Modelling inhibition, where the occurrence of an event reduces the intensity of future arrival, has drawn relatively less attention in the literature.  However, inhibitory effects are prevalent in various domains. For example, in neuroscience, inhibitory kernel functions can represent the presence of a latency period before the successive activations of a neuron \citep{reynaud2013inference}.
	
	This paper proposes a non-parametric estimation model for the non-linear Hawkes process. The non-linear Hawkes process allows the inclusion of both excitatory and inhibitory effects to model a  broader range of phenomena. The stability condition of the nonlinear Hawkes process is explored in \cite{bremaud1996stability}.  \citet{bonnet2021maximum} use a negative exponential function to model inhibitive kernels for a univariate nonlinear Hawkes process and use maximum likelihood estimation to determine the optimal parameters.  \cite{bonnet2022inference} extend this approach to a multivariate nonlinear Hawkes process. \cite{lemonnier2014nonparametric} develop the Markovian Estimation of Mutually Interacting Process (MEMIP) method, which utilizes weighted exponential functions to determine kernels for the nonlinear Hawkes process. To extend the MEMIP for large dimensional datasets, \cite{lemonnier2016multivariate} introduce dimensionality reduction features. However, the above approaches require the kernel function to be smooth, which is a drawback. \cite{wang2016isotonic} propose an algorithm that learns the nonlinear Hawkes kernel non-parametrically using isotonic regression and a single-index model. The Isotonic Hawkes Process, however, assumes only a continuous monotonic decreasing excitation kernel to capture the influence of the past arrivals. 
	
	In their work, \cite{du2016recurrent} employ a recurrent neural network (RNN) to model event timings and markers simultaneously by leveraging historical data. Conversely, \cite{mei2017} propose a novel continuous-time long short-term memory (LSTM) model to capture the self-modulating Hawkes processes, capable of accounting for the inhibiting and exciting effects of prior events on future occurrences. Moreover, this approach can accommodate negative background intensity values, which correspond to delayed response or inertia of some events. While RNN-based models can capture complex long-term dependencies, inferring causal relationships or extracting causal information from these models can be challenging. 
	
	We propose a neural networks based non-parametric model for the nonlinear Hawkes process where feed-forward networks are used to model the kernels and time-varying base intensity functions. The architecture of the neural network is chosen such that the likelihood function and its gradients with respect to the network parameters can be efficiently evaluated. As the likelihood function is non-convex with respect to the parameter space, we use the Stochastic Gradient Descent (SGD) with Adam \citep{kingma2014adam} to obtain the optimal network parameters that maximize the log-likelihood. The method is an extension of the Shallow Neural Hawkes model proposed by \cite{joseph2022shallow}, which was designed for linear Hawkes processes and only allows for the modelling of kernels with an excitation feature. We evaluate our model against state-of-the-art methods for non-linear Hawkes processes, using both simulated and real datasets.

	The paper is structured as follows: Section \ref{Sec:prelim} provides a definition of the non-linear Hawkes process and formulates the associated log-likelihood maximization problem. In Section \ref{Sec:model}, we introduce the proposed neural network model for the non-linear Hawkes process and discuss the parameter estimation procedure. Section \ref{Sec:results} presents the results obtained by our method on synthetic data and its application to infer the order dynamics of bitcoin and ethereum market orders on the Binance exchange. Additionally, we apply our method to a high-dimensional neuron spike dataset. Finally, in Section \ref{Sec:conclusion}, we summarize our findings and discuss the limitations of our method.

	\section{Preliminary Definitions}\label{Sec:prelim}

	\paragraph{Definition of the Hawkes Process:}
	We denote a D-dimensional counting process as $N(T),$ and its associated discrete events  as $\mathcal{S} = (t_n,d_n)_{n \ge 1}$, where $t_n \in [0,T)$ is the timestamp of the $n$th event and $d_n \in (1,2,...,D)$ is the dimension in which the $n$th event has occurred. Let  $\{t_1^d, \ldots,t_k^d,\ldots, t^d_{N_d(T)} \},$ be the ordered arrivals for the dimensions $d=1,\ldots,D$. Given  $t \ge 0$, the count of events in $[0,t)$ for the $d$th dimension will be $$N_d(t)= \sum_{n \ge 1} \mathbbm{1} _{t^d_n < t}.$$  The conditional intensity function for the counting process at $t,$ $\lambda^*_d(t): \R^+ \rightarrow \R^+,$ is given by,
	
	\begin{equation}
		\label{Eq:PPintensity}
		\lambda^*_d(t) =\underset{\Delta t \rightarrow 0}{lim} \frac{{\mathbb{E}\left[N_d(t+\Delta t )-N_d(t)\right | \mathcal{H}_{t^-} ]}}{\Delta t} 
	\end{equation}
	
	where, $\mathcal{H}_{t^-},$ denotes the history of the counting process upto $t.$
	
	For a $D$-dimensional Hawkes process the conditional intensity $\lambda_d(t),$ for the $d$-th dimension is expressed as,
	
	\begin{equation}
		\lambda_d(t)=\mu_d(t) + \sum_{j=1}^{D} \sum_{\{\forall k |(t_k^j<t)\}} \phi_{dj}(t-t_k^j),
		\label{Eq:multidim_Hawkes}
	\end{equation}
	
	where $\mu_d(t),$ the exogenous base intensity for the $d$-th dimension, does not depend upon the history of the past arrivals. $\phi_{dj}(t-t_k^j),$  $1\leq d,j\leq D,$ are called the excitation kernels that quantify the magnitude of excitation of the conditional intensity for the $d$-th dimension at time $t$ due to the past arrival at $t_k^j,$ $\{\forall k | t_k^j < t\}$ in the $j$th dimension. These kernel functions are positive and causal, i.e., their support is in $\mathbb{R}^+.$ Inferring a Hawkes process requires estimating the base intensity function $\mu_d$ and its kernels functions $\phi_{dj},$ either by assuming a parametric form for the kernels or in a non-parametric fashion. A more generic form of the Hawkes process conditional intensity ${\lambda}^*_d(t)$, is given by,
	\begin{equation}
		\label{Eq:approInte}
		{\lambda}^*_d(t) = \Psi_d({\lambda}_d(t)), \textrm{   } \forall d=1,...,D,
	\end{equation}
	where $\Psi_d : \R \rightarrow \R^+.$
	
	If $\Psi_d$ is an identity function then, Equation \ref{Eq:approInte} is equivalent to the linear Hawkes process expressed in Equation \ref{Eq:multidim_Hawkes}. However, if $\Psi_d$ is a non-linear function, then the process is called a Non-linear Hawkes Process. For the stability and stationarity of the nonlinear Hawkes process,  $\Psi_d$ should be Lipschitz continuous, as given by Theorem 7 of \cite{bremaud1996stability}.  The advantage of the nonlinear Hawkes process is that it allows the kernel output to take negative values to model inhibitory effects as well as positive values to model excitatory effects, i.e.,  $\phi_{dj}: \R^+ \rightarrow \R.$  \citet{wang2016isotonic} uses a sigmoid function $(1+e^{-x})^{-1}$ and a decreasing function $1-(1+e^{-x})^{-1},$ while \citet{bonnet2021maximum} and \citet{costa2020renewal} use a $\max$ function as $\Psi.$

	\paragraph{The Log-Likelihood Function of non-linear Hawkes Process}
	
	The categorization of methods used for estimating Hawkes processes, as discussed in \cite{cartea2021gradient}, can be broadly classified into three groups:
	\begin{itemize}
		\item \emph{Maximum likelihood estimation (MLE)} It is the most commonly used approach for estimating the kernels of the Hawkes process, as employed by many methods such as \cite{bonnet2021maximum}, \cite{du2016recurrent}, and \cite{lemonnier2016multivariate}. However, MLE methods have high computational costs as their time complexity increases quadratically with the number of arrivals. Expectation maximization methods, such as those used by \cite{lewis2011nonparametric} and \cite{zhou2013learning}, also belong to this category.
		
		\item \emph{Method of moments} \cite{wang2016isotonic} and \cite{achab2017uncovering} use a moment matching method to estimate the Hawkes process. These methods typically rely on the spectral properties of the Hawkes process. The WH method \citep{bacry2014second}, which is used as a benchmark model in numerical experiments, estimates the Hawkes process as a system of Wiener-Hopf equations and also falls under this category.
		
		\item \emph{Least squares estimation (LSE)} This method involves minimizing the L2 error of the integrated kernel. \cite{cartea2021gradient} propose an efficient stochastic gradient-based estimation method for linear multivariate Hawkes processes, which applies to large datasets and general kernels. However, the method is not applicable suitable for nonlinear Hawkes process. 
		
	\end{itemize}	
	
	In this paper the optimal kernels are estimated through the maximum likelihood approach.

	For a  $D$-multidimensional nonlinear Hawkes process with $\boldsymbol{\mu}=[\mu_d(t)]_{D \times 1}$ and $\boldsymbol{\Phi}=[\phi_{dj}(t)]_{D \times D}$ ; $1 \le d,j \le D$, let $\boldsymbol{\theta}=[\theta_1,\ldots\theta_p \ldots \theta_P]$ be the set of parameters used to model $\boldsymbol{\mu}$ and $\boldsymbol{\Phi}$.  The parameters can be estimated by maximizing the log-likelihood function over the sampled events from the process. The log-likelihood (LL) function corresponding to the dataset $\mathcal{S}$ for the nonlinear Hawkes process is given by (see for instance \cite{rubin1972regular},\cite{daley2007introduction}),
	\begin{equation}
		\label{Eq:LL}
		\mathcal{L}(\boldsymbol{\theta},\mathcal{S}) = \sum_{d=1}^{D} \left[ \sum_{t_n,d_n\in \mathcal{S}} \left\{\log(\lambda^*_d(t_n))- \int_{t^d_{n-1}}^{t^d_n} \lambda^*_d(s) ds\right\} \mathbbm{1} _{d_n = d}\right]
	\end{equation}

	Depending on the parametric form of the kernel,$\phi_{dj}(t),$  the LL function may or may not be concave. For instance, even for the exponential kernel, $\phi_{dj}(t) =\alpha_{dj}\exp(-\beta_{dj})(t),$ the LL function is not concave in the parameter space. We use the Stochastic Gradient Descent (SGD) method to search the local optima for the LL function in the parameter space. In order to estimate the parameters $\boldsymbol{\theta}$ using the SGD we need an unbiased estimator of the gradient of the LL, as given by Equation \ref{Eq:LL},  with respect to $\theta.$
	
	\begin{eqnarray}\nonumber
		\nabla_{\theta_p} \left(\mathcal{L}(\boldsymbol{\theta},\mathcal{S}) \right)&=& \nabla_{\theta_p} \left( \sum_{d=1}^{D} \left[ \sum_{t_n,d_n\in \mathcal{S}} \left\{\log(\lambda^*_d(t_n))- \int_{t^d_{n-1}}^{t^d_n} \lambda^*_d(s) ds\right\} \mathbbm{1} _{d_n = d} \right] \right)\\
		&=&\sum_{d=1}^{D} \left[ \sum_{t_n,d_n\in \mathcal{S}} \nabla_{\theta_p} \left(\left\{\log(\lambda^*_d(t_n))- \int_{t^d_{n-1}}^{t^d_n} \lambda^*_d(s) ds\right\}\right) \mathbbm{1} _{d_n = d}\right]\label{Eq:LLUnbiased}
	\end{eqnarray}
	
	Following Equation \ref{Eq:LLUnbiased},  an unbiased estimator of the gradient of $\mathcal{L}$ with respect to parameter $\theta_p$  will be,
	
	\begin{equation}
		\label{Eq:LLGrad}
		\nabla_{\theta_p}\widehat{\mathcal{L}}(\boldsymbol{\theta},t^d_n):=\nabla_{\theta_p}\left(\log(\lambda^*_d(t^d_n))- \int_{t^d_{n-1}}^{t^d_n} \left(\lambda^*_{d}(s)\right) ds \right),
	\end{equation}
	where $t^d_n$ is randomly sampled from $\mathcal{S}.$ The proposed model does not assume a specific parametric form for the kernels and the base intensities. The model and its estimation is discussed in  Section \ref{Sec:model}. 
	
	\section{Proposed Model}\label{Sec:model}
	
	\cite{hornik1989multilayer} show that a multilayer feed-forward networks with as few as one hidden layer are capable of universal approximation in a very precise and satisfactory sense.  We here propose a neural network-based system called the Neural Network for Non-Linear Hawkes (NNNH) to estimate the kernels and the base intensity of the nonlinear Hawkes process. Each kernel, $\phi_{dj},\,1\leq d,j \leq D,$ and base intensity function $\mu_d,$ is separately modelled as a feed-forward network. However, the parameters of all the networks are jointly estimated by maximizing the log-likelihood of the training dataset.

	In the NNNH, each kernel $\phi_{dj}$, $1 \le d,j \le D$  is modelled as a feed-forward neural network $\hat{\phi}_{dj}: \R^+ \rightarrow \R, $
	
	\begin{equation}
		\label{Eq:Model}
		\widehat{\phi}_{dj}(t)=A_2 \circ \varphi \circ A_1,
	\end{equation}
	
	where $A_1$: $\mathbb{R}$ $\rightarrow$ $\mathbb{R}^p$ and $A_2$:  $\mathbb{R}^p$ $\rightarrow$ $\mathbb{R}.$ More precisely, 
	\begin{equation*}
		A_1(x) = \mathbf{W}_1 x + \mathbf{b_1} \ \ \textrm{for} \ x \in \R,\ \mathbf{W}_1 \in \R^{p \times 1}, \mathbf{b_1} \in \R^p,
	\end{equation*}
	\begin{equation*}
		A_2(\mathbf{x}) = \mathbf{W}_2\mathbf{x} +b_2  \ \ \textrm{for} \ \mathbf{x} \in \R^p,\ \mathbf{W}_2 \in \R^{1 \times p}, b_2	 \in \R,
	\end{equation*}
	and $\varphi : \R^j \rightarrow \R^j, j \in \mathbb{N}$ is the component-wise Rectified Linear Unit (ReLU) activation function given by:
	\begin{equation*}
		\varphi(x_1,\ldots,x_j):=\left(\max(x_1,0),\ldots,\max(x_j,0)\right).
	\end{equation*}

	We take $\mathbf{W}_1 =[a^1_1,a^2_1,....,a^p_1]^{\top}$, $\mathbf{W}_2 =[a^1_2,a^2_2,....,a^p_2]$ and $\mathbf{b_1} =[b^1_1,b^2_1,....,b^p_1]^{\top}.$ Therefore, the kernel function can be written as:
	\begin{equation}
		\label{Eq:NNequation}
		\widehat{\phi}_{dj}(x) = b_2+ \sum_{i=1}^{p}a^i_2  \max\left(a^i_1 x+b^i_1,0\right) 
	\end{equation}
	
	As the kernel, $\widehat{\phi}_{dj},$ maps from positive real numbers to real numbers, it is versatile in its ability to model both inhibitory and excitatory effects.
	
	\begin{figure}[H]
		\centering
		\begin{tikzpicture}
			
			\node (hidden_mu1)[linear1]  at (0,-0.5) {$\varphi$};
			\node (hidden_mu2)[linear1, yshift=-2.5cm] {$\varphi$};
			\node (hidden_mu3)[linear1, yshift=-5.0cm] {$\varphi$};
			
			\node[box_weight, left of = hidden_mu1,xshift = -0.5cm ] (a_11) {$\sum$};
			\node[bias, below of = a_11 ] (b1_11) {};
			\node[box_weight, left of = hidden_mu2,xshift = -0.5cm ] (a_12) {$\sum$};
			\node[bias, below of = a_12] (b1_12){};
			\node[box_weight, left of = hidden_mu3,xshift = -0.5cm ] (a_1p) {$\sum$};
			\node[bias, below of = a_1p ] (b1_1p){};

			\node[input, left of = a_12, yshift = -0.1cm, xshift = -1cm ] (input) {$x$};

			\node[box_weight, right of = hidden_mu2,xshift = 2cm ] (a_22) {$\sum$};
			\node[bias, below of = a_22,yshift=0.0cm ] (b2_22) {};
			
			\node (linear_mu) [linear1,right of = a_22,xshift = 1cm]{$f$};
			\node[input, right of = linear_mu, xshift = 1.5cm ] (output) {};
			\draw [dashed] (hidden_mu2) -- (hidden_mu3);
			
			\draw [arrow] (input) -- (a_11) node[midway,above]  {$a^1_1$} ;
			\draw [arrow] (input) -- (a_12) node[midway,above]  {$a^2_1$};
			\draw [arrow] (input) -- (a_1p) node[midway,above]  {$a^p_1$};
			
			\draw [arrow] (a_11) -- (hidden_mu1);
			\draw [arrow] (a_12) -- (hidden_mu2);
			\draw [arrow] (a_1p) -- (hidden_mu3);
			
			\draw [arrow] (hidden_mu1) -- (a_22) node[pos=0.6,above]  {$a^1_2$} node[pos=0.2, above]  {$g^1_1$};
			\draw [arrow] (hidden_mu2) -- (a_22) node[pos=0.6,above]  {$a^2_2$} node[pos=0.3,above left]  {$g^2_1$};
			\draw [arrow] (hidden_mu3) -- (a_22) node[pos=0.5,above right]  {$a^p_2$} node[pos=0.3,above left]  {$g^p_1$};
			
			\draw [black] (a_11) -- (b1_11) node[midway,right=2pt]  {$b^1_1$};
			\draw [black] (a_12) -- (b1_12) node[midway,right=2pt]  {$b^2_1$};
			\draw [black] (a_1p) -- (b1_1p) node[midway,right=2pt]  {$b^p_1$};
			
			\draw [arrow] (a_22) -- (linear_mu) node[midway,above]  {$g_2$};

			\draw [black] (a_22) -- (b2_22) node[midway,right=2pt]  {$b_2$};

			\draw [arrow] (linear_mu) -- (output) node[midway,above]  {$\widehat{\phi}_{dj}(x)$};
			\node[rectangle,draw,dotted,minimum width = 4.0cm, minimum height = 6.4cm, fill=white, fill opacity=0.2] (NN) at (-0.8,-3.1) {};
			\node [input, above of =  hidden_mu1, xshift=-0.7cm]  () {\small Hidden Layer};
			\node [input, above of = a_22, xshift=0.7cm, yshift=2cm]  () {\small Output Layer};
			\node [input, below of =  hidden_mu2, xshift=-0.7cm, yshift=-3.2cm]  () {\small $g_1=\varphi(a_1x+b_1)$};
			\node [input, below of =  a_22, xshift=0.7cm, yshift=-3.2cm]  () {\small $g_2=f(a_2g_1+b_2)$};
			\node [input, below of =  hidden_mu2,  xshift=1.5cm, yshift=-4cm]  () {\small $\widehat{\phi}_{dj}(x) = f\left(b_2+a_2\varphi\left(a_1x+b_1\right)\right)$};
		\end{tikzpicture}
		\caption{The architecture of the feed-forward neural network used for modeling kernels and base intensity functions of the nonlinear Hawkes process in NNNH. Here $\varphi,$ $f$ are the activation functions. In the NNNH we take $\varphi$ as ReLU and  $f$ as the identity function.}\label{Fig:kernelModel}
	\end{figure}
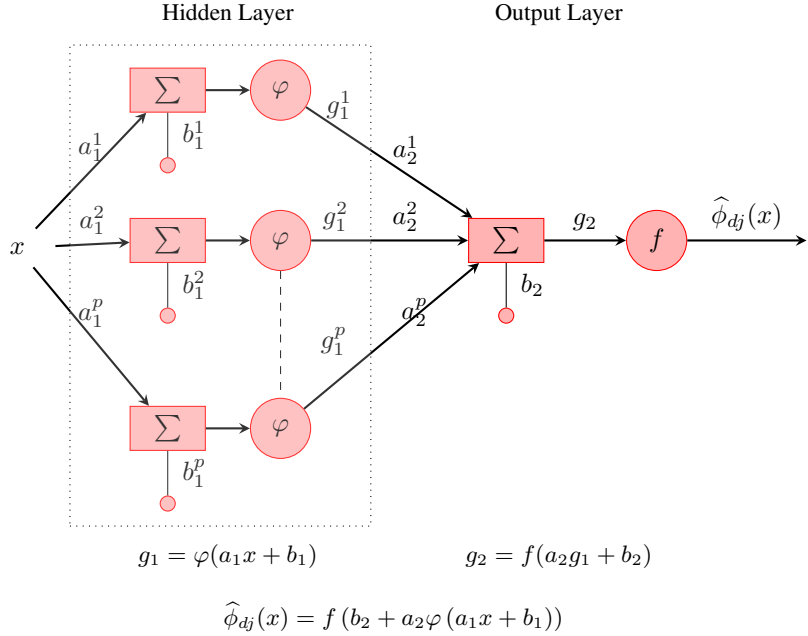

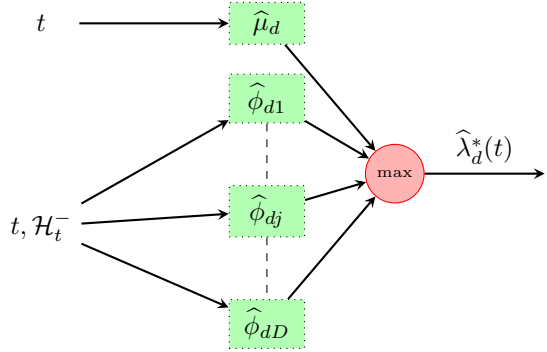
\begin{figure}[H]
		\centering
		\begin{tikzpicture}
			
			\node[box] (mu) at (0,5) {$\widehat{\mu}_d$};
			\node[box, below of = mu ] (phid1) {$\widehat{\phi}_{d1}$};
			\node[box, below of = phid1,yshift = -0.5cm ] (phidd) {$\widehat{\phi}_{dj}$};
			\draw [dashed] (phid1) -- (phidd);
			\node[box, below of = phidd,yshift = -0.5cm ] (phidD) {$\widehat{\phi}_{dD}$};
			\draw [dashed] (phidd) -- (phidD);
			\node[linear, right of = phidd,yshift = 0.5cm, xshift = 0.7 cm](total) {\tiny $\max$};
			\draw [arrow] (mu) -- (total);
			\draw [arrow] (phid1) -- (total);
			\draw [arrow] (phidd) -- (total);
			\draw [arrow] (phidD) -- (total);
			\node[input, left of = phidd,yshift = -0.2cm, xshift = -2cm ] (input) {$t,\mathcal{H}_t^-$};
			\draw [arrow] (input) -- (phid1) ;
			\draw [arrow] (input) -- (phidd);
			\draw [arrow] (input) -- (phidD);
			\node[input, left of = mu, xshift = -2cm ] (inputMu) {$t$};
			\draw [arrow] (inputMu) -- (mu);
			\node[input, right of = total,xshift = 1.5cm] (output) {};
			\draw [arrow] (total) -- (output) node[midway,above]{$\widehat{\lambda}^*_d(t)$};
			
		\end{tikzpicture}
		\caption{NNNH model for multivariate nonlinear Hawkes process.}
		\label{Fig:model}
	\end{figure}

	With a choice of $p$ neurons for the hidden layer, the dimension of the parameter space for the above network will be $3p+1.$ For a $D$-dimensional nonlinear Hawkes process there will be $D^2$ kernels and the total number of parameters to be estimated would be $(3p+1)D^2.$ In case the base intensity is not constant, we also model it as a feed-forward neural network, with an architecture similar to the ones used for the kernels. We then need to approximate in total $(3p+1)\left(D^2+D\right)$ parameters. If the base intensity is assumed to be a constant, the total number of parameters to be estimated would be $(3p+1)D^2+ D.$ The  Figure \ref{Fig:model} is a schematic representation of the NNNH model, with each $\phi_{dj},$ and $\mu_d$ being a network (as expressed in Figure \ref{Fig:kernelModel}). In order to estimate $\lambda_d^*(t),$ we need in addition to $t,$ the history of arrivals until $t,$ i.e., $\mathcal{H}_t^-.$ The NNNH therefore estimates $\lambda^*_d(t)$ as:
	
	\begin{equation}
		\widehat{\lambda}^*_d(t)=\max\left(\widehat{\mu}_d(t) + \sum_{j=1}^{D} \sum_{\{\forall k |(t_k^j<t)\}} \widehat{\phi}_{dj}(t-t_k^j),0\right).
		\label{Eq:multidim_HawkesApprox}
	\end{equation}
	
	The outer $\max$ function ensures that the estimated $\lambda^*_d(t)$ is positive and mimics the $\Psi$ function in the non-linear Hawkes process, as given in Equation \ref{Eq:approInte}.
	
	This non-parametric method employs a feed-forward neural network that can precisely estimate any continuous kernel function and base intensity function defined on a compact set, with a desired level of accuracy (\cite{leshno1993multilayer}). However, the challenge is to jointly estimate the parameters of the $D^2 + D $ neural networks used to model the base intensity and the kernel functions. To do this, the set of parameter values that maximize the log-likelihood over the observed dataset $\mathcal{S}$ is found. Since the log-likelihood function is non-convex in the parameter space, the batch stochastic gradient descent is used to estimate the parameter values that provide a local maxima of the log-likelihood function The gradient of the log-likelihood function as given by Equation \ref{Eq:LLGrad} with respect to each network parameter is computed for updating the parameter values in the batch stochastic gradient method.
	
	\subsection{Estimating the parameters of the NNNH}\label{Sec:gradient}
	
	The challenge for the NNNH model is, given the dataset $\mathcal{S}$ to efficiently obtain the parameters of the networks that locally maximize the log-likelihood function. More precisely, let $\boldsymbol{\theta}=[\theta_1,\ldots\theta_p \ldots \theta_P]^{\top}$ denote the parameters of the network and let $\boldsymbol{\Theta}$ denote the parameter space; then we want to find, for given dataset $\mathcal{S},$ the values of the parameters that maximize the log-likelihood function over the parameter space, i.e.,
	\begin{equation}
		\widehat{\boldsymbol{\theta}} = \argmax_{\boldsymbol{\theta} \in \boldsymbol{\Theta}} \mathcal{L}\left(\boldsymbol{\theta}, \mathcal{S}\right).
	\end{equation}
	Gradient descent is an effective optimization technique when the log-likelihood function can be differentiated with respect to its parameters. This is because computing the first-order partial derivatives of all parameters has the same computational complexity as evaluating the log-likelihood function, making it a relatively efficient approach. With neural networks, stochastic gradient descent, has been shown to be more effective and efficient method for optimization. The NNNH parameter estimation algorithm utilizes Adam (\cite{kingma2014adam}), a stochastic gradient descent technique that adjusts learning rates adaptively and relies solely on first-order gradients. Equation \ref{Eq:LLGrad} provides the necessary unbiased estimator of the log-likelihood function for the NNNH method to be used with the SGD. Algorithm \ref{alg:deep-learning-gradient-descent} gives the pseudo-code for the NNNH parameter estimation using Adam.

	\begin{algorithm}[H]
		\SetAlgoLined
		\KwIn{ dataset $\mathcal{S}$ }
		\KwIn{learning rate $\alpha =0.01$, decay rates $\beta_1=0.9, \beta_2=0.999$, small constant $\epsilon=10^{-8}$}
		\KwOut{Optimal parameter value $\widehat{\boldsymbol{\theta}}$}
		initialize parameter vector $\boldsymbol{\theta_0}$;\\
		$m_0 \gets 0$ (initialize first moment vector);\\
		$v_0 \gets 0$ (initialize second moment vector);\\
		initialize iteration step $i = 0$;\\
		\While{stopping criterion not met}{
			$i \gets i + 1$\\
			randomly sample $(t_i,d_i) \in \mathcal{S}$ \\
			$g_i \gets \nabla_{\boldsymbol{\theta}} \widehat{\mathcal{L}}(\boldsymbol{\theta}_{i-1},t_i^d)$ compute unbiased gradient of the likelihood function \\
			$m_i \leftarrow \beta_1 m_{i-1} + (1 - \beta_1) g_i$  update biased first moment estimate \\
			$v_i \leftarrow \beta_2 v_{i-1} + (1 - \beta_2) g_i^2$   update biased second moment estimate \\
			$\hat{m}_i \leftarrow \frac{m_i}{1 - \beta_1^i}$  compute bias-corrected first moment estimate \\
			$\hat{v}_i \leftarrow \frac{v_i}{1 - \beta_2^i}$  compute bias-corrected second moment estimate \\
			$\boldsymbol{\theta}_i \leftarrow \boldsymbol{\theta}_{i-1} - \alpha \frac{\hat{m}_i}{\sqrt{\hat{v}_i} + \epsilon}$  update parameters 
		}
		\caption{Estimation of parameters of NNNH using Adam} \label{alg:deep-learning-gradient-descent}
	\end{algorithm}
	
	The essential step while estimating the parameters for the NNNH model is computing the unbiased gradient of the likelihood function, i.e.,
	
	\begin{equation*}
		\nabla_{\theta_p}\widehat{\mathcal{L}}(\boldsymbol{\theta},t^d_n):=\nabla_{\theta_p}\left(\log(\widehat{\lambda}^*_d(t^d_n))- \int_{t^d_{n-1}}^{t^d_n} \left(\widehat{\lambda}^*_{d}(s)\right) ds \right).
	\end{equation*}
	
	This in turn involves computing the following gradients,
	
	$$
	\nabla_{\theta_p}\log(\widehat{\lambda}^*_d(t^d_n)) \, \text{and} \, \nabla_{\theta_p} \int_{t^d_{n-1}}^{t^d_n} \left(\widehat{\lambda}^*_{d}(s)\right) ds.
	$$
	
	While it is possible to compute the former gradient efficiently, the challenge lies in accurately estimating the latter gradient. The former gradient calculation involves the following steps:
	
	\begin{eqnarray}\nonumber
		\nabla_{\theta_p}\log(\widehat{\lambda}^*_d(t^d_n)) &=& \frac{1}{\widehat{\lambda}^*(t^d_n)} \nabla_{\theta_p} \max\left(\widehat{\mu}_d(t_n^d) + \sum_{j=1}^{D} \sum_{\{\forall k |(t_k^j<t_n^d)\}} \widehat{\phi}_{dj}(t_n^d-t_k^j),0\right), \\\nonumber
		&=&\frac{1}{\widehat{\lambda}^*(t^d_n)} \left(\nabla_{\theta_p} \left(\widehat{\mu}_d(t_n^d)\right) + \sum_{j=1}^{D} \sum_{\{\forall k |(t_k^j<t_n^d)\}} \nabla_{\theta_p} \left( \widehat{\phi}_{dj}(t_n^d-t_k^j)\right)\right)\mathbbm{1}_{\widehat{\lambda}_d(t_n^d)>0}.
	\end{eqnarray}

	Here, 
	\begin{equation*}
		\widehat{\lambda}^*_d(t)=\max\left(\widehat{\lambda}_d(t),0\right).
	\end{equation*}
	
	In Appendix \ref{App:gradKernel}, you can find the expressions for the gradients of the neural networks $\widehat{\mu}_d(t)$ and $\widehat{\phi}_{dj}(t).$ In Section \ref{Subsec:IntLambda} we provide an approach for efficient estimation of $\nabla_{\theta_p} \int_{t^d_{n-1}}^{t^d_n} \left(\widehat{\lambda}^*_{d}(s)\right) ds.$

	\subsection{Gradient of the integrated Hawkes intensity function}\label{Subsec:IntLambda}
	
	The computation of the first order derivatives of,
	$$
	\widehat{\Lambda}_d(t_n^d) := \int_{t^d_{n-1}}^{t^d_n} \widehat{\lambda}^*_{d}(s) ds,
	$$	
	with respect to all the parameters will have the same computational complexity as evaluating the function itself. In the SGD method, it is crucial to perform efficient computations of $\widehat{\Lambda}_d(t_n^d)$ since these gradients must be computed multiple times.  Gaussian Quadrature is a numerical technique that can be used to estimate the definite integral, but it is limited to well-behaved integrands that can be approximated by a certain degree of polynomial. Unfortunately, the non-linear Hawkes intensity function lacks smoothness, making the use of Gaussian Quadrature prone to substantial error. Moreover, Gaussian Quadrature entails computing the quadrature points and weights, which can be time-consuming and resource-intensive. The number of quadrature points required for accurate approximation also grows quickly with the polynomial degree, resulting in increased computational cost.
	
	Our selection of network architecture can enable efficient evaluation of the integral $\widehat{\Lambda}_d(t_n^d)$ in the NNNH. The above integral can be expanded as:
	\begin{eqnarray}\nonumber
		\int_{t^d_{n-1}}^{t^d_n} \widehat{\lambda}^*_{d}(s) ds  &=& \int_{t^d_{n-1}}^{t^d_n} \max\left(\widehat{\lambda}_{d}(s),0\right) ds \\\nonumber
		&=& \int_{t^d_{n-1}}^{t^d_n} \max \left(  \left[ \widehat{\mu}_d(s) + \sum_{j=1}^{D} \sum_{\{t_k^j<s\}} \widehat{\phi}_{dj}(s-t_k^j)\right],0       \right) ds\\\nonumber
		&=&\int_{t^d_{n-1}}^{t^d_n} \max \left( \left[  \widehat{\mu} + \sum_{j=1}^{D} \sum_{\{t_k^j<s\}} \sum_{i=1}^{p} a^i_2  \max\left(a^i_1 (s-t_k^j)+b^i_1,0\right)\right] , 0 \right) ds,
	\end{eqnarray}
	
	with $\widehat{\phi}_{dj}(s)$ substituted using Equation \ref{Eq:NNequation} and  $\widehat{\mu}_d(s)$ set as a constant for simplicity in the third line. In addition, we set the bias $b_2$ to zero for ease of exposition. Therefore, the above integral takes the following form:
	
	\begin{equation}
		\int_{t^d_{n-1}}^{t^d_n} \widehat{\lambda}^*_{d}(s) ds = \int_{t^d_{n-1}}^{t^d_n} \max\left( \left[\widehat{\mu} +\sum_{\{k,i,j\}} a^i_2  \max\left(a^i_1 (s-t_k^j)+b^i_1,0\right)\right],0\right) ds
	\end{equation}
	
	Let $\mathbf{X} \equiv \{x_1,\ldots,x_{P^*}\}$ be the sequence of \emph{zero-crossings} obtained for all the $P^*$ neurons, where the zero-crossing for the $i$th neuron is obtained by solving:
	\begin{eqnarray}\\\nonumber
		a^i_1 (x_i-t_k^j)+b^i_1 &=& 0\\\nonumber
		x_i &=& t_k^j - \frac{b^i_1}{a^i_1}.
	\end{eqnarray}
	
	$P^*$ represents the total number of neurons used in the neural networks for modeling the kernels, $\widehat{\phi}_{dj}$ with $1 \leq d, j \leq D, $ of a $D$-dimensional nonlinear Hawkes process.  For example, if each kernel in a $D$-dimensional nonlinear Hawkes process is modelled using a network with $P$ neurons, then $P^* = P(D^2)$.

	We re-index the sequence $\mathbf{X} $ in monotone increasing order as $\mathbf{S} \equiv \{s_1 \leq s_2 \leq \cdots \leq s_{P^*} \}.$ Let $\mathbf{S}^* \equiv \{s_l \leq \cdots \leq s_{u} \},$ where $t_{n-1}^d \leq s_l\leq\cdots\leq s_u\leq t_n^d,$ and $1\leq l\leq u\leq P^*,$
	be the largest subsequence of the sorted sequence $\mathbf{S}.$ Therefore, $\mathbf{S}^*$ is the set of all zero-crossings that lie in the range $[t_{n-1}^d, t_n^d].$ 
	
	We then write,
	$$
	\int_{t^d_{n-1}}^{t^d_n} \widehat{\lambda}^*_{d}(s) ds = \int_{t^d_{n-1}}^{s_l} \max\left(\widehat{\lambda}_d(s), 0\right) ds+\ldots\int_{s_{q-1}}^{s_q} \max\left(\widehat{\lambda}_d(s), 0\right) ds  \ldots+ \int_{s_u}^{t_n^d} \max\left(\widehat{\lambda}_d(s), 0\right) ds,
	$$
	
	and exploit the fact that $\widehat{\lambda}_d(s)$ will be linear in $s$ in the sub-intervals, $[t_{n-1}^d,s_l],\ldots,[s_u, t_n^d].$ If $\widehat{\lambda}_d(s)$ is linear in $s$ in the interval $[s_{q-1},s_q],$ then there will be at the max one zero crossing for the function $\max(\widehat{\lambda}_d(s),0)$ that lies within this interval. If $s^{\text{out}}_q$ is the zero-crossing for $\max(\widehat{\lambda}_d(s),0)$ and $s_q \leq s^{\text{out}}_q \leq s_q,$ we can write: 
	$$
	\int_{s_{q-1}}^{s_q} \max\left(\widehat{\lambda}_d(s), 0\right) ds = \left(\int_{s_{q-1}}^{s^{\text{out}}_q} \widehat{\lambda}_d(s) ds\right) \mathbbm{1}_{\widehat{\lambda}_d(s_{q-1}) >0} + \left(\int_{s^{\text{out}}_q}^{s_q} \widehat{\lambda}_d(s) ds\right)\mathbbm{1}_{\widehat{\lambda}_d(s_{q}) >0}.
	$$
	
	In case $s^{\text{out}}_q \notin [s_{q-1},s_q]$ the above can be integrated as:
	$$
	\int_{s_{q-1}}^{s_q} \max\left(\widehat{\lambda}_d(s), 0\right) ds = \left(\int_{s_{q-1}}^{s_q} \widehat{\lambda}_d(s) ds\right)\mathbbm{1}_{\widehat{\lambda}_d(s_{q-1}) >0}.
	$$
	
	Therefore, we evaluate $\widehat{\Lambda}_d(t_n^d),$ by splitting the integral into intervals within which $\max(\widehat{\lambda}_d(s),0)$ is linear in $s.$ The integral,
	$$
	\int_{t^d_{n-1}}^{t^d_n}\widehat{\lambda}^*_{d}(s)ds
	$$
	as well as its gradients with respect to the network parameters can then be exactly evaluated. The expression of the above integral in a sub-interval $[s_{q-1},s_q],$ where $\widehat{\lambda}^*_d(s)$ is linear is provided in the Appendix \ref{App:integralExpression}.

	\section{Experiments and Results} \label{Sec:results}
	
	In this section, we evaluate the effectiveness of the NNNH method on synthetic and real-world datasets\footnote{The source code and the dataset used in the experiments described here are available at
		https://github.com/sobin-joseph/NNNH}.. We assess the performance of the NNNH method in estimating the conditional intensity function of a non-homogeneous Poisson Processes (NHPP). We then showcase the adaptability of the NNNH method by utilizing one-dimensional non-linear Hawkes processes. Specifically, we explore the NNNH's capacity to estimate Hawkes processes with smooth, non-smooth, and negative kernels, and examine its robustness in handling different variants of non-linear Hawkes models. Additionally, we investigate the NNNH's ability to estimate kernels and base intensity function for multidimensional non-linear Hawkes processes, where we evaluate a simultaneous combination of smooth, non-smooth, and inhibitive kernels.

	We investigate the practical applications of non-linear Hawkes models through two case studies. In the first case, we apply the NNNH method to analyze tick-by-tick cryptocurrency trading data on the Binance exchange, seeking to identify any causal relationships between the buy and sell market orders of the BTC/USD and the ETH/USD pairs. For the second case, we utilize the NNNH to analyze neuronal spike trains recorded from the motor cortex of a monkey, revealing the interdependence between individual neurons and how they function in tandem.
	
	Prior to delving into the outcomes of our numerical experiments, we provide a brief overview of the data preprocessing steps and the hyper-parameters selected for fitting the NNNH model to a dataset. Specifically we discuss the scaling approach adopted for the dataset, the choice of the number of neurons opted for the network, the initial learning rates employed for the Adam optimizer, and the stopping criteria for the optimizer. 
	
	\subsection{Data preprocessing and choice of hyper-parameters:}
	
	For all our experiments, we divide the dataset into training, validation, and test set.  The dataset is partitioned in a 60:20:20 ratio for analysis.  Due to the dependence on history in the Hawkes process, we do not alter the chronology of the events. Before splitting the dataset, we first scale the timestamps. Scaling the dataset is critical step in the preprocessing stage of building neural networks as it can enhance their performance, stability, and convergence. Additionally, scaling helps to ensure that all input features have a similar range of variance, which can lead to improved initialization of the network parameters and overall performance during training. Given dataset $\mathcal{S}= (t_n,d_n),$ where $t_n \in [0,T),$ let  $\{t_1^d, \ldots,t_k^d,\ldots, t^d_{N_d(T)} \},$ be the ordered arrivals for the dimensions $d=1,\ldots,D$. We define $T_{\max}$ as
	$$
	T_{\max} = \max\left(t^1_{N_1(T)}, \ldots, t^d_{N_d(T)}, \ldots, t^D_{N_D(T)}\right),
	$$
	
	and $N(T_{\max})$ as:
	$$
	N(T_{\max})= \sum_{d=1}^D N_d(T_{\max}).
	$$ 
	
	We scale the original timestamps as: 
	$$
	\widehat{t}_n^d= t_n^d \frac{N(T_{\max})}{T_{\max}}.
	$$

	\paragraph{Initialization of network parameters and choice of batch size}
	
	In the NNNH method as discussed in Section \ref{Sec:model} we model each kernel and base intensity function as a feed-forward neural network. To model the base intensity functions, $\widehat{\mu}_d(t),$ we use a feed-forward network with fifty neurons in the hidden layer. The following initialization is used for the parameters of the network, 
	
	\begin{eqnarray}\nonumber
		a_1^i &\sim& \mathcal{U}(-10^{-3} , 10^{-3})\\\nonumber
		a_2^i &\sim& \mathcal{U}(0 , 0.2)\\\nonumber
		b_1^i &\sim& \mathcal{U}(-1 , 1),
	\end{eqnarray}
	
	where $\mathcal{U}(a , b),$ denotes uniform distribution between $a$ and $b.$ 
	
	For the kernels, $\widehat{\phi}_{dj}(t),\, 1 \leq d,j \leq D,$  of the nonlinear Hawkes process we use feed-forward networks with thirty two neurons in the hidden layer and the following scheme for initialising the network parameters,
	
	\begin{eqnarray}\nonumber
		a_1^i &\sim& \mathcal{U}(0 , -0.3)\\\nonumber
		a_2^i &\sim& \mathcal{U}(0 , 0.2)\\\nonumber
		b_1^i &\sim& \mathcal{U}(0 , 0.3).
	\end{eqnarray}
	
	If we use a constant baseline intensity we initialize $\widehat{\mu} = 1.$ 
	
	We use a batch size of hundred for calculating the stochastic gradient. We use different learning rates for updating the parameters in the hidden layer and the output layer, as we observe faster convergence with this choice. The following learning rates were used for all the experiments:
	For fitting the networks used to model the base intensity function we used a learning rate of $10^{-3}$ for updating the network parameters in the output layer and $10^{-6}$ for updating the parameters in the hidden layer.  For the networks used to model the kernel function the we used corresponding learning rates of $10^{-2}$ and $10^{-3}$ for the output and hidden layers, respectively.
	
	\paragraph{Stopping criteria:}
	
\begin{figure}[h!]
		
		\centering
		\includegraphics[height=0.3\textheight,width=\textwidth]{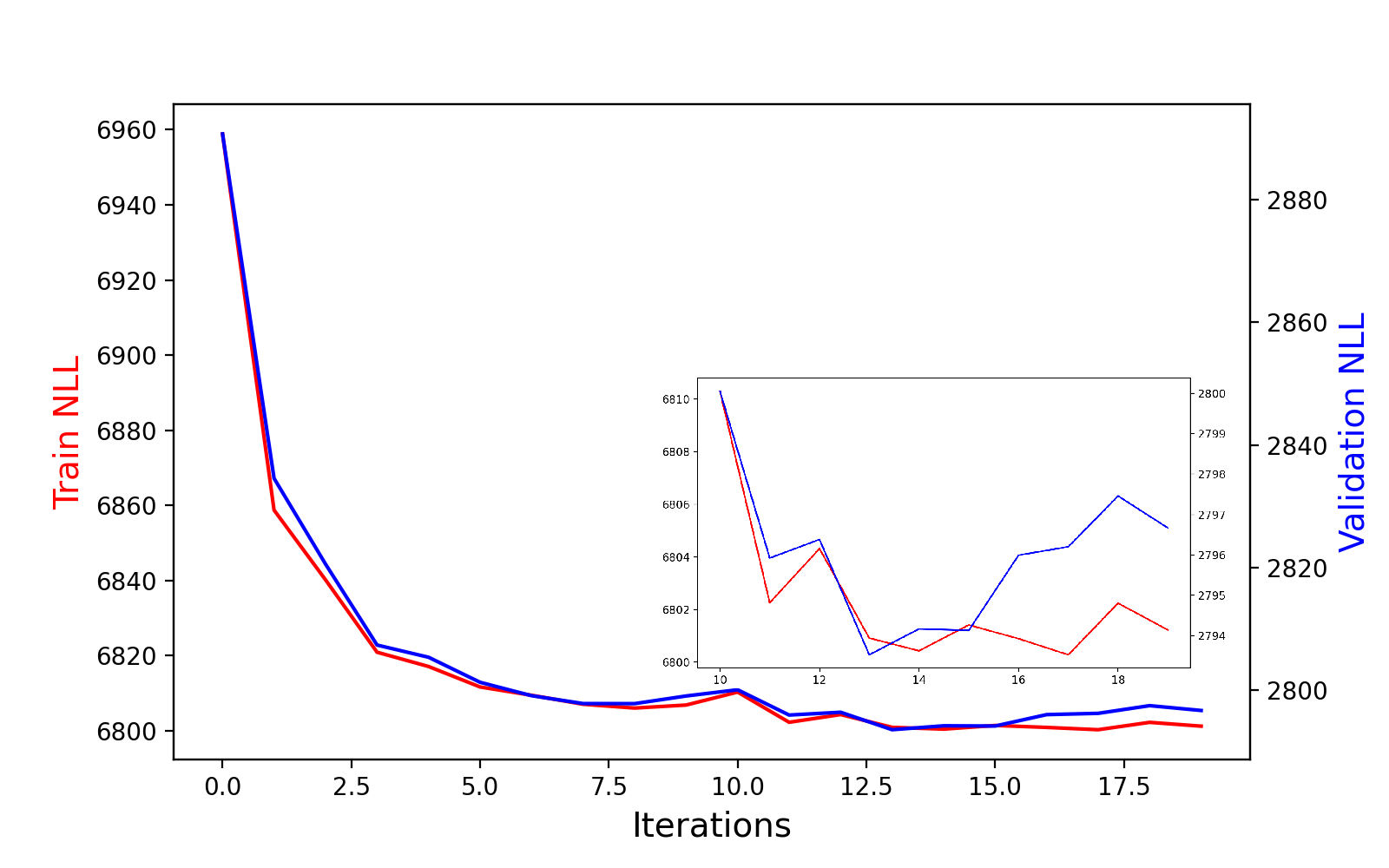}
		
		\caption{Plot of Negative log-likelihood for train and validation datasets in one dimensional case,  indicating the stopping criteria. }
		\label{Fig:1dTraValErr}
	\end{figure}

	In this study, for all the experiments, we utilize a stopping criterion based on the negative log-likelihood value computed on the validation dataset. The negative log-likelihood values are computed at the end of each iteration of the batch stochastic gradient descent algorithm. The NNNH parameter estimation algorithm stops when the updated parameters fail to improve the best recorded validation error for a specified number of iterations, following the approach proposed by \citet{goodfellow2016deep}. This early stopping criterion helps prevent over fitting, as demonstrated in Figure \ref{Fig:1dTraValErr}, where we observe a reduction in training negative log-likelihood value while the validation negative log-likelihood value starts increasing after a certain number of iterations, indicating that the model is starting to over-fit the training data.
	
	Through numerical experiments, Section \ref{Sec:sens} presents our investigation of the sensitivity of NNNH estimation to various hyper-parameter choices.

	\subsection{Estimation of Non Homogenous Poisson Process Intensity}\label{Subsec:EstimationNHP}

	A Non-Homogenous poisson Process (NHPP) $N(t),t\geq0$ is a generalized counting process in which the rate at which events occurs varies with time. The intensity function, denoted by $\mu(t): \mathbb{R}^+ \rightarrow \mathbb{R}^+$, captures the varying rate of the events over time, which could be influenced by factors such as external events, seasonal patterns, or other underlying phenomena. To comprehend this counting process, estimating the intensity function is necessary. The intensity function of NHPP can be estimated parametrically or non-parametrically. Parametric estimation assumes that the parametric form of the underlying intensity function is known. For instance, \cite{rigdon1989power} estimates the parameters by assuming the intensity function as power law function, while \cite{lee1991modeling} uses a general exponential-polynomial-trigonometric intensity function. If one does not assume the parameteric form of the intensity function, they would resort to non-parametric techniques to estimate the conditional intensity function. In \cite{leemis1991nonparametric}, a technique is proposed that utilizes a piecewise linear function to estimate the cumulative intensity of the NHPP, while \citet{xiao2013wavelet} presents another non-parametric approach that employs a wavelet-based estimation method to estimate the intensity function.
	
	In this section, we show that the NNNH technique can be used to model the intensity function, $\mu(t),$ of a non-homogeneous Poisson process. A single-layered feed-forward neural network can be used to model the intensity function as,
	
	\begin{equation}\label{Eq:muModel}
		\hat{\mu}(t)= \max\left(b_2+\sum_{i=1}^{p}a^i_2  \max(a^i_1 t+b^i_1,0), 0 \right),
	\end{equation}
	
	where, as explained in Section\ref{Sec:model}, $[a^i_1, b^i_1, a^i_2, b_2],\, 1 \leq i \leq p,$ are the  parameters of the network that are estimated by maximizing the log-likelihood value,
	
	\begin{equation}\label{Eq:NHPLL}
		\mathcal{L}(\boldsymbol{\Theta})= \sum_{t_n \in \mathcal{S}} log(\widehat{\mu}(t_n)) - \int_{0}^{T}\widehat{\mu}(s) ds,
	\end{equation}	
	
	for the training dataset $\mathcal{S}.$ We use the SGD with Adam, for adaptive learning rates, to find the optimal network parameters, as discussed in Section \ref{Sec:gradient}.
	
	We consider an exponential, linear, polynomial, and a trigonometric function, as given in Table \ref{tab1}, to model the conditional intensity. The arrival times are simulated using the thinning algorithm of \citet{lewis1979simulation}. Table \ref{tab1} compares the true negative log-likelihood of the simulated NHPP with the negative log-likelihood values obtained using the fitted NNNH model \footnote{The source code and the dataset used in the experiments described here are available at
		https://github.com/sobin-joseph/NNNH}.
	
	\begin{table}[h] 
		\centering
		\caption{The NHPP functions and their corresponding True Negative Log-Likelihood(TNLL) and NNNH Negative Log-Likelihood(NNLL) obtained from the NNNH method }\label{tab1}
		\begin{tabular}{ |c|l|c|c| } 
			\hline
			Function & Underlying equation & TNLL & NNLL \\
			\hline \hline
			Exponential & $\mu(t)=0.5e^{-0.001t}$ & 1177&1173\\
			\hline 
			Linear & $\mu(t)=0.5$&2528&2518 \\
			\hline
			Parabola & $\mu(t)=2 \times 10^{-7}(1.5t-2000)$ & 1703 &1699\\
			\hline 
			Sin Curve & $\mu(t)=0.4(\textrm{sin}(2\pi t \times0.0004-1000)+1.1)$ & 3482 &3538\\
			\hline
		\end{tabular}
		
		\label{Tab:PPNLL}
	\end{table}
	
\begin{figure}[h!]
		
		\centering
		\subcaptionbox{Exponential}{\includegraphics[width=0.45\textwidth,height=0.2\textheight]{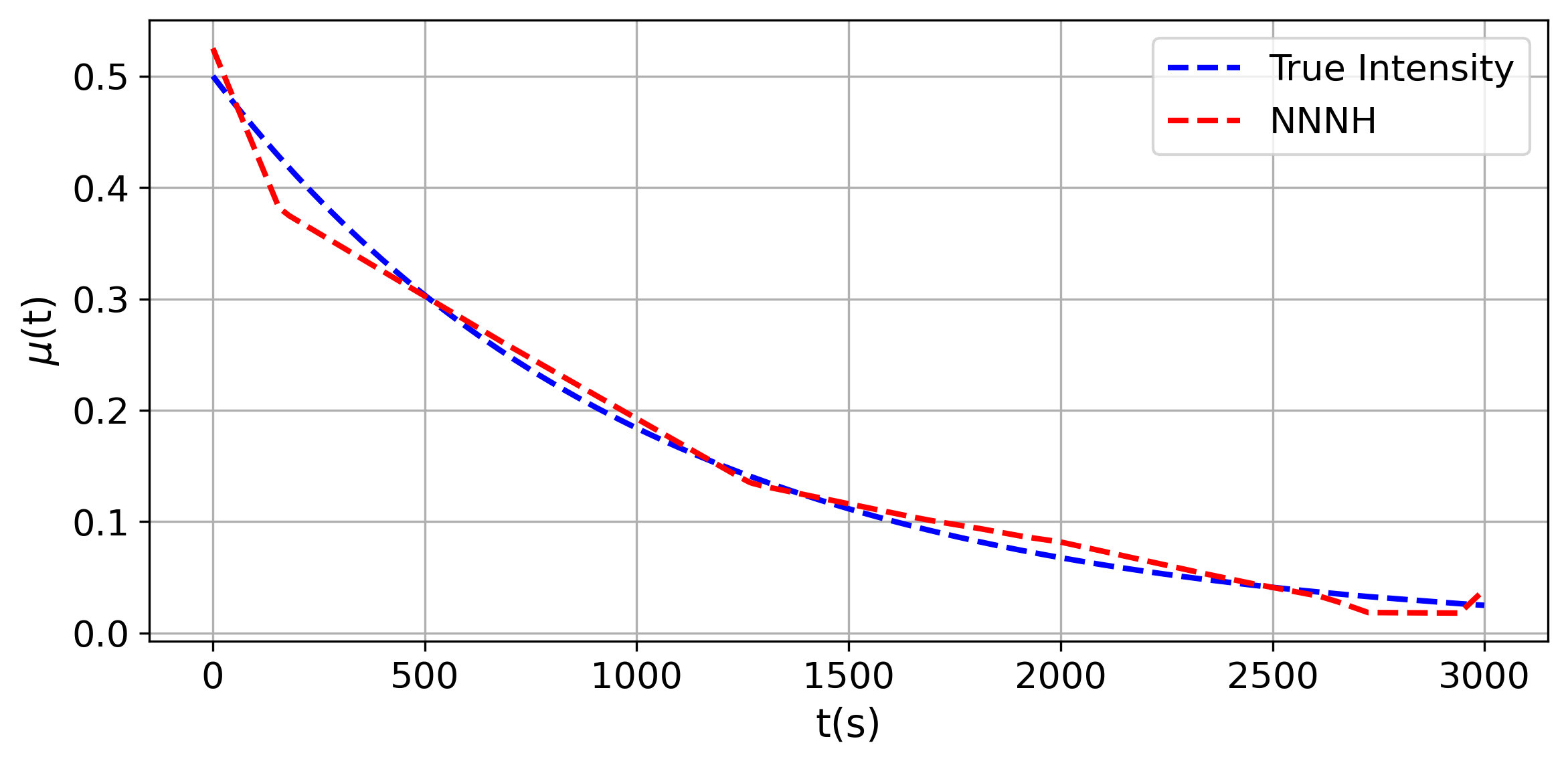}}%
		\hfill
		\subcaptionbox{Parabola}{\includegraphics[width=0.45\textwidth,height=0.2\textheight]{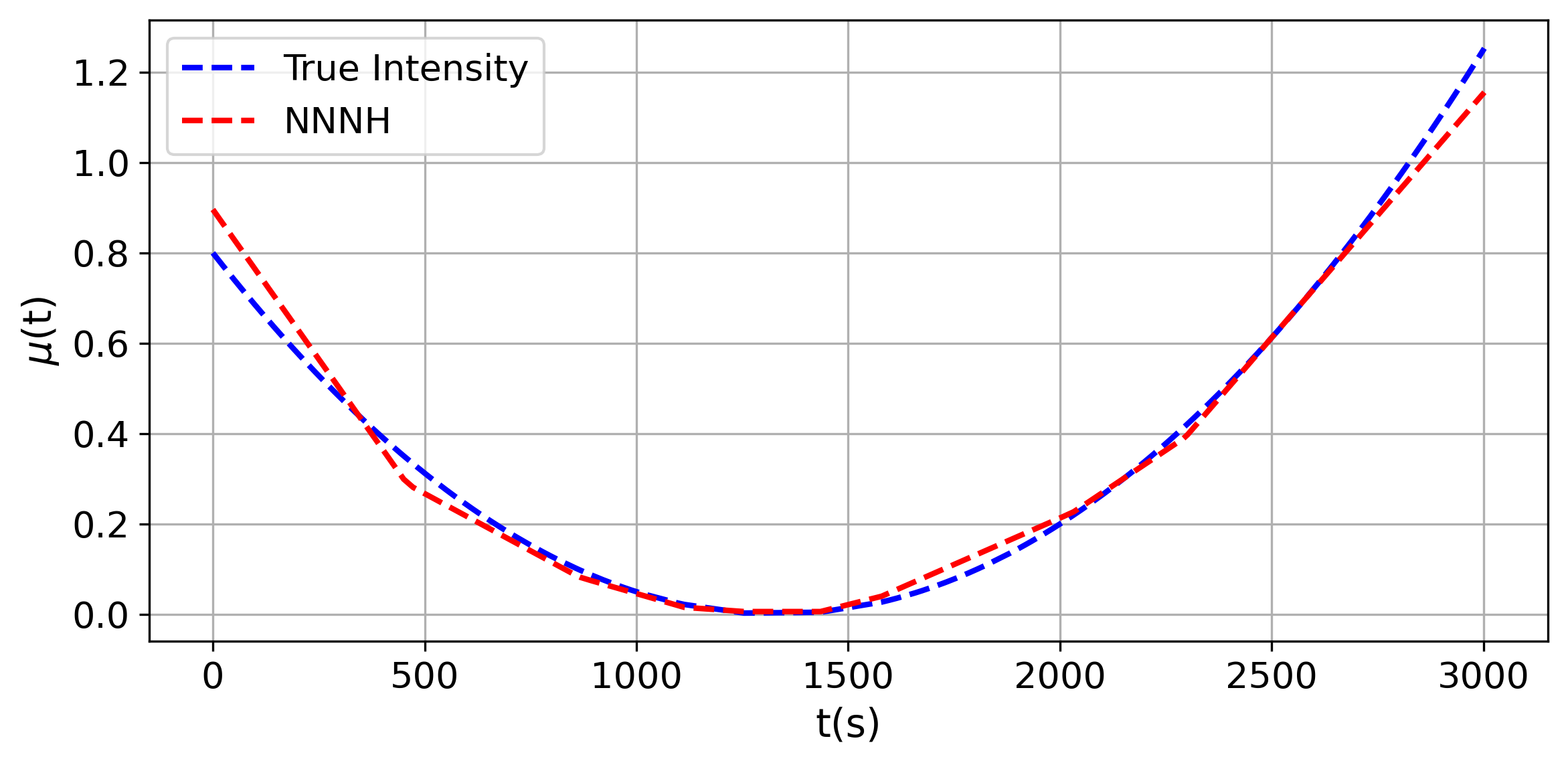}}%
		\hfill 
		\subcaptionbox{constant}{\includegraphics[width=0.45\textwidth,height=0.2\textheight]{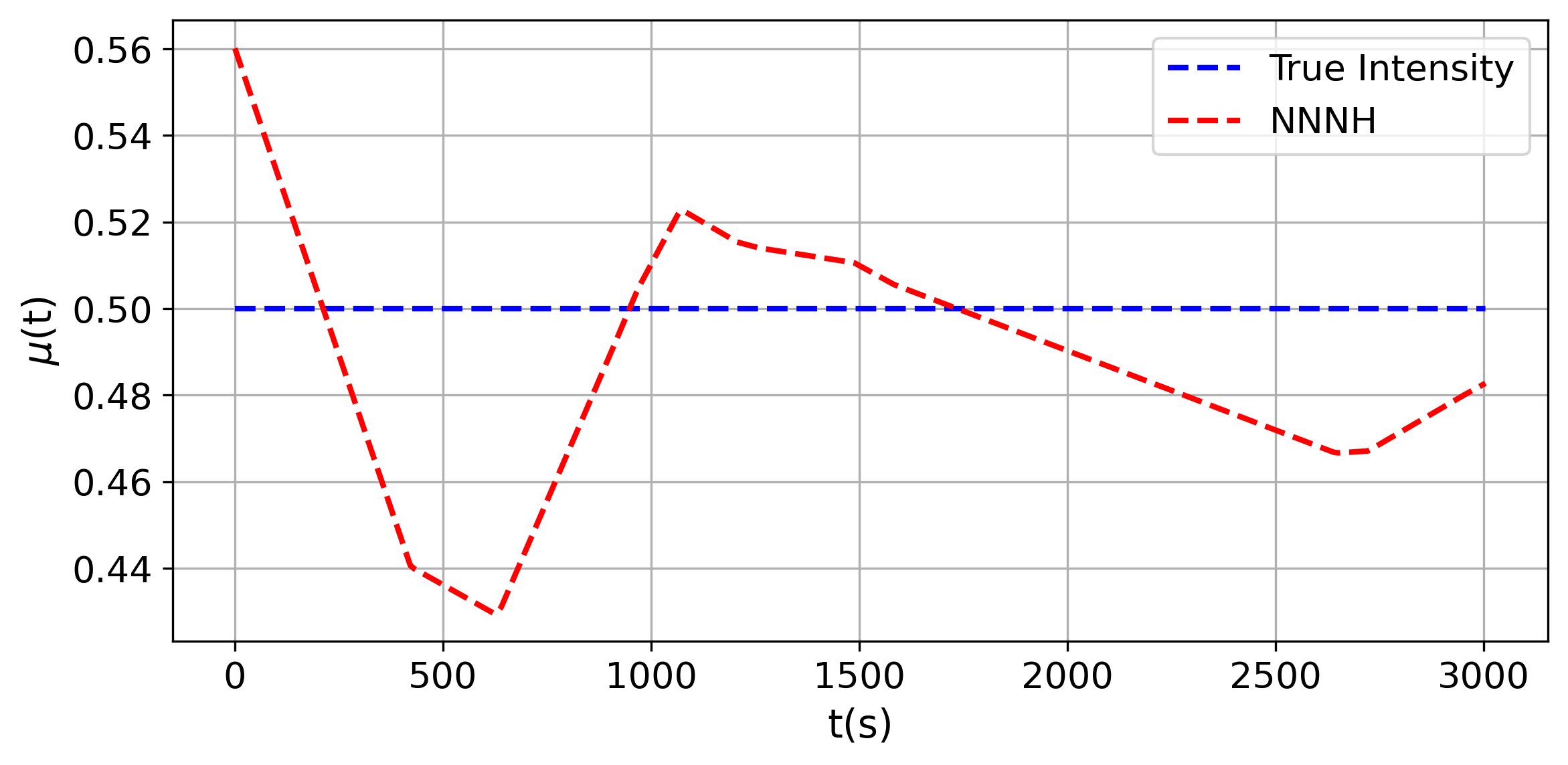}}%
		\hfill 
		\subcaptionbox{Sine}{\includegraphics[width=0.45\textwidth,height=0.2\textheight]{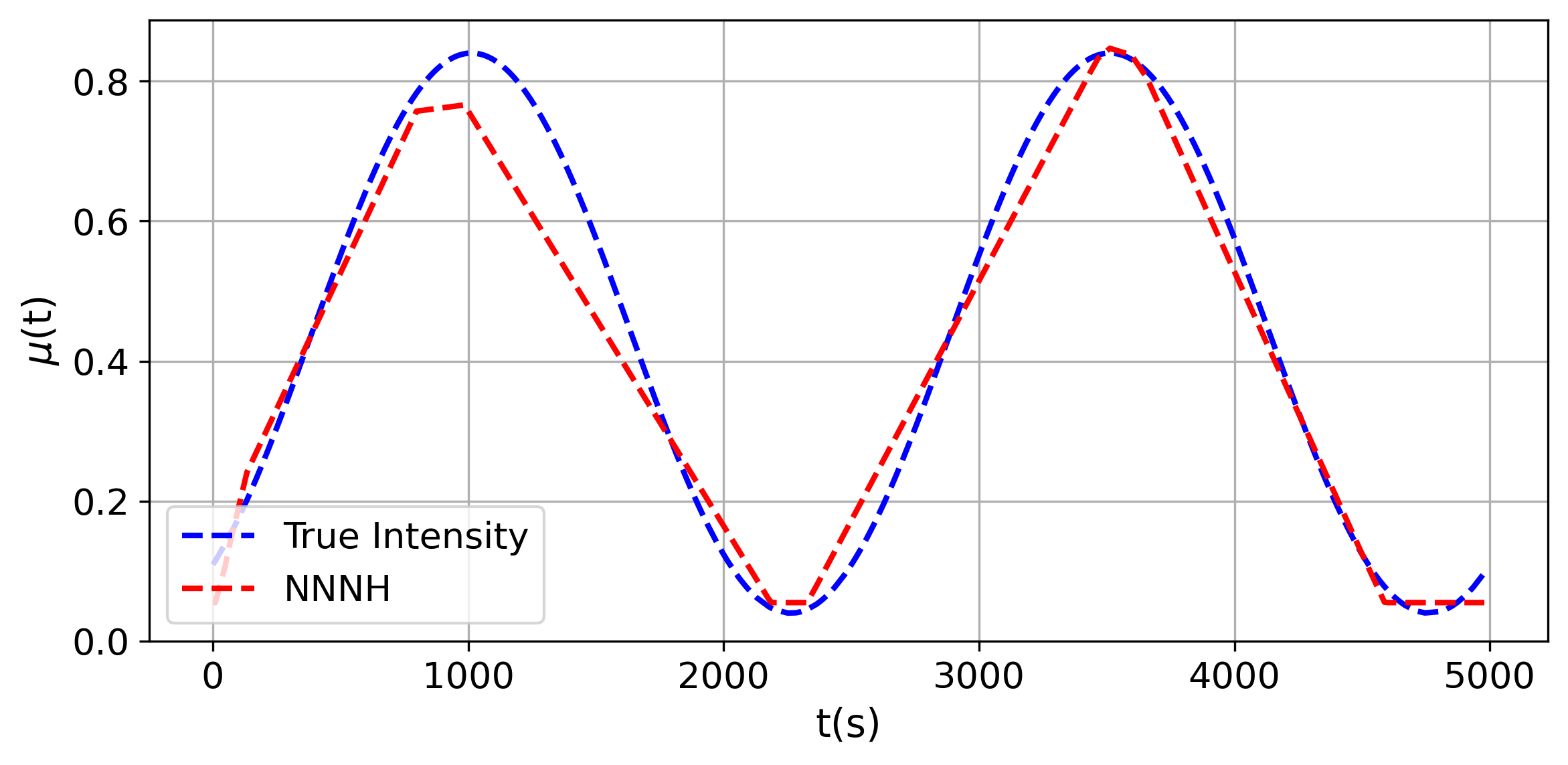}}%
		\caption{NNNH estimated kernel and base intensity for a one-Dimensional Hawkes process with sin base intensity and exponential kernel}
		
		\label{Fig:PPEstim}
	\end{figure}

	We see that for all the cases considered, the estimated negative log-likelihood values are reasonably close to the true negative log-likelihood values. 
	Figure \ref{Fig:PPEstim} compares the NNNH estimates of the intensity function with the true intensity function. We see for all the cases considered the NNNH is able to recover the intensity function reasonably well. 

	\subsection{Univariate Hawkes Estimation}
	\label{Subsec:UniEst}
	
	This section focuses on examining the effectiveness of the NNNH method in estimating one-dimensional non-linear Hawkes processes using the following criteria: 
	\begin{itemize}
		\item Estimation  of non-smooth kernels
		\item Estimation of negative kernels
		\item Estimation of non-linear Hawkes processes for different variations of $\Psi.$
		\item Estimation of Hawkes processes with varying base intensity function. 
	\end{itemize} 
	
	We simulate the arrival times for the different variants of the Hawkes processes using the \textit{Ogata's thinning algorithm} proposed in \citet{ogata1981lewis}.
	We compare the performance of our algorithm to the following state of the art methods: 
	
	\begin{itemize}
		\item WH: An algorithm proposed in \citet{bacry2014second}, a non-parametric estimation method which solves a Wiener–Hopf system derived from the auto-covariance of the multivariate Hawkes processes. 
		\item Bonnet: A maximum likelihood based estimation method for Hawkes processes with self excitation or inhibition as proposed in  \citet{bonnet2021maximum}. 
	\end{itemize}

	\paragraph{Estimation of non-smooth kernels:} 
	
	We first consider a linear Hawkes process with a non-smooth kernel, specifically a rectangular kernel of the form,
	
	$$\textrm{Rectangular,\,\,\,} 
	\phi(t) = 
	\left\{
	\begin{array}{l}
		\alpha\beta, \textrm{if } \delta \le t \le \delta+\frac{1}{\beta}, \\
		0, \textrm{otherwise}
	\end{array}
	\right.$$
	
	The smoothness assumption of the kernel is a prerequisite for some non-parametric methods used in estimating the kernels of Hawkes processes, such as the Markovian Estimation of Mutually Interacting Process (MEMIP) proposed in \cite{lemonnier2014nonparametric}. As a result, it is necessary to compare the performance of the NNNH method using a non-smooth kernel in order to assess its effectiveness.
	
	We simulate the process using the following parameter values for the rectangular kernel,
	
	$$
	\mu =0.05,\, \alpha = 0.7, \, \beta = 0.4, \, \delta = 1.
	$$
	
\begin{figure}[H]
		
		\centering
		\includegraphics[width=0.80\textwidth]{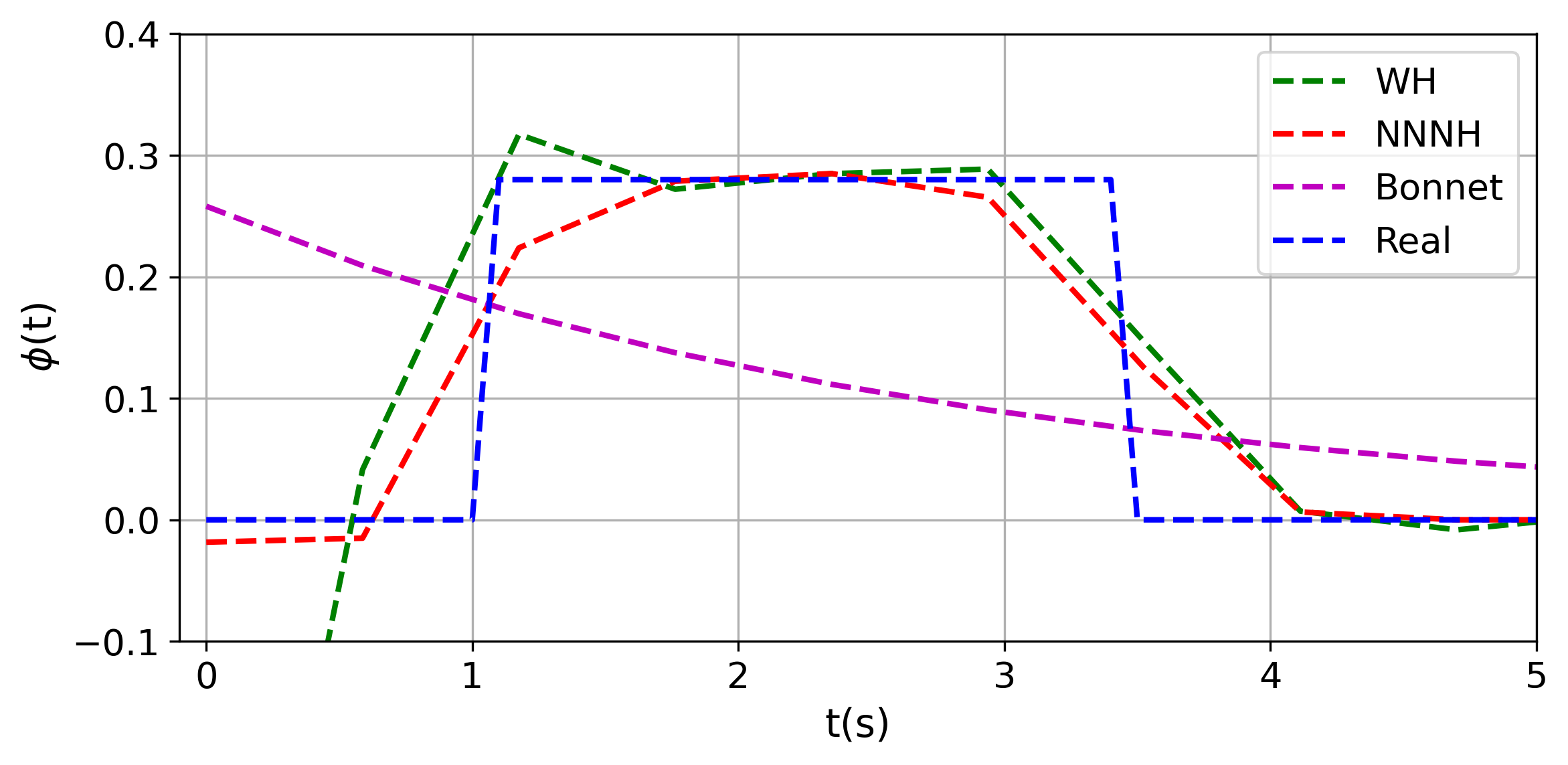}%
		
		\caption{Fitted Rectangular kernel: A comparison of the NNNH, WH, and Bonnet estimates with the actual (real) kernel. }
		\label{Fig:1aRect}
	\end{figure}

	Simulating the process until $T = 60000$ yields $N(T) = 10001$ arrivals. As a first step, we visually compare the estimated kernels obtained using the NNNH and the benchmark methods with the true kernel. Figure \ref{Fig:1aRect} illustrates the kernels fitted by the three estimation methods. As expected we see that the non-parametric models, i.e., the NNNH and the WH are able to better recover the ground truth. This is also reflected in the negative log-likelihood values obtained for WH, Bonnet, and NNNH methods, i.e., $4228$, $12838$, and $2800$ respectively. 
	
	\paragraph{Estimation of negative kernels:}
	
	We next consider a non-linear Hawkes process, 
	$$\lambda^*(t) = \max\left(\mu + \sum_{\tau<t_n} \phi(t-\tau), 0\right),$$
	where, $\Psi$ in Equation \ref{Eq:approInte} is a $\max$ function, i.e. $\Psi(\lambda(t)) = \max(\lambda(t),0).$ The inhibitive kernel, $\phi,$ is specified as an exponential function given by,
	$$\phi(t) = -0.5 e^{-2t }.$$ 
	
	Given that non-parametric methods, such as the Expectation Maximization (EM) approach introduced in \cite{lewis2011nonparametric}, are capable of estimating only positive kernels, it is of interest to evaluate the capacity of the NNNH method in estimating the negative kernels.
	
\begin{figure}[H]
		
		\centering
		
		\includegraphics[width=0.80\textwidth]{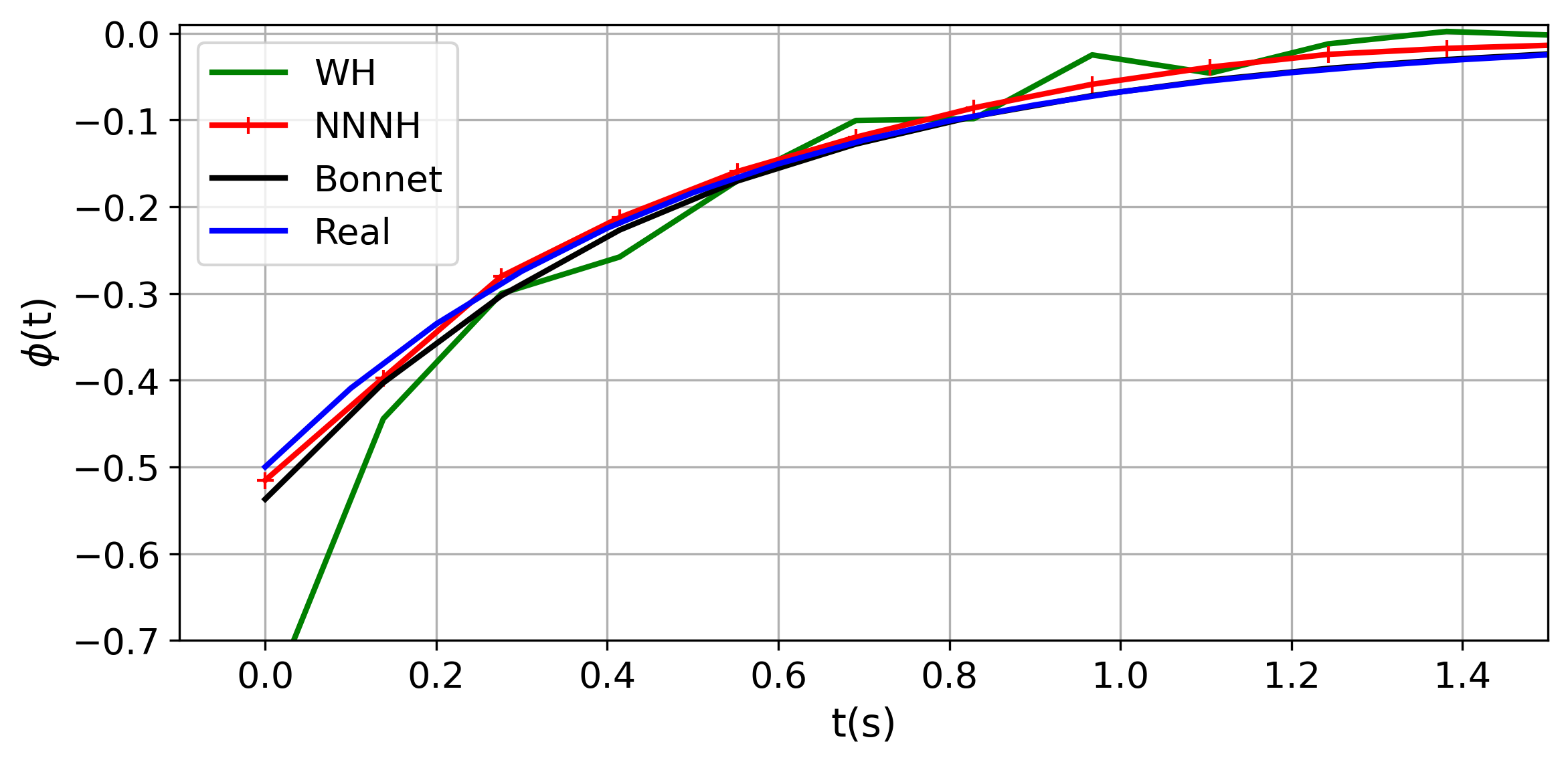}%
		\caption{Fitted negative exponential kernel: A comparison of teh NNNH, WH, and Bonnet estimates with the actual kernel. }
		\label{Fig:1dConsMu}
	\end{figure}

	We simulate the above process for $T = 14000,$ with a base intensity, $\mu =0.9.$ This resulted in $N(T) =10001$ events. We first compare the kernel obtained using the  NNNH, WH, and Bonnet method with the ground truth in Figure \ref{Fig:1dConsMu}.  All three methods are able to recover the kernel, but visually the NNNH and Bonnet methods appear to provide better estimates. Furthermore, this conclusion is supported by the estimated negative log-likelihood values, which are $9833,$ $9559,$ and $9564$ for the WH, Bonnet, and the NNNH, respectively.

	\paragraph{Estimation of non-linear Hawkes processes for different variations of $\Psi:$}
	
	Diverse variations of the non-linear Hawkes process can be derived by making distinct selections of the function $\Psi$ in Equation \ref{Eq:approInte}. In contrast to the preceding example where $\Psi$ was chosen as a $\max$ function, in this instance, we adopt a sigmoid function for $\Psi$ given by
	
	$$\lambda^*(t) = \frac{1}{1+e^{-(\lambda(t)-2)}}$$.
	We consider the following kernel, 
	$$
	\phi(t) =  e^{-2 t },$$
	
	for this example. 
	
\begin{figure}[h!]
		\includegraphics[width= \textwidth, height =0.2\textheight]{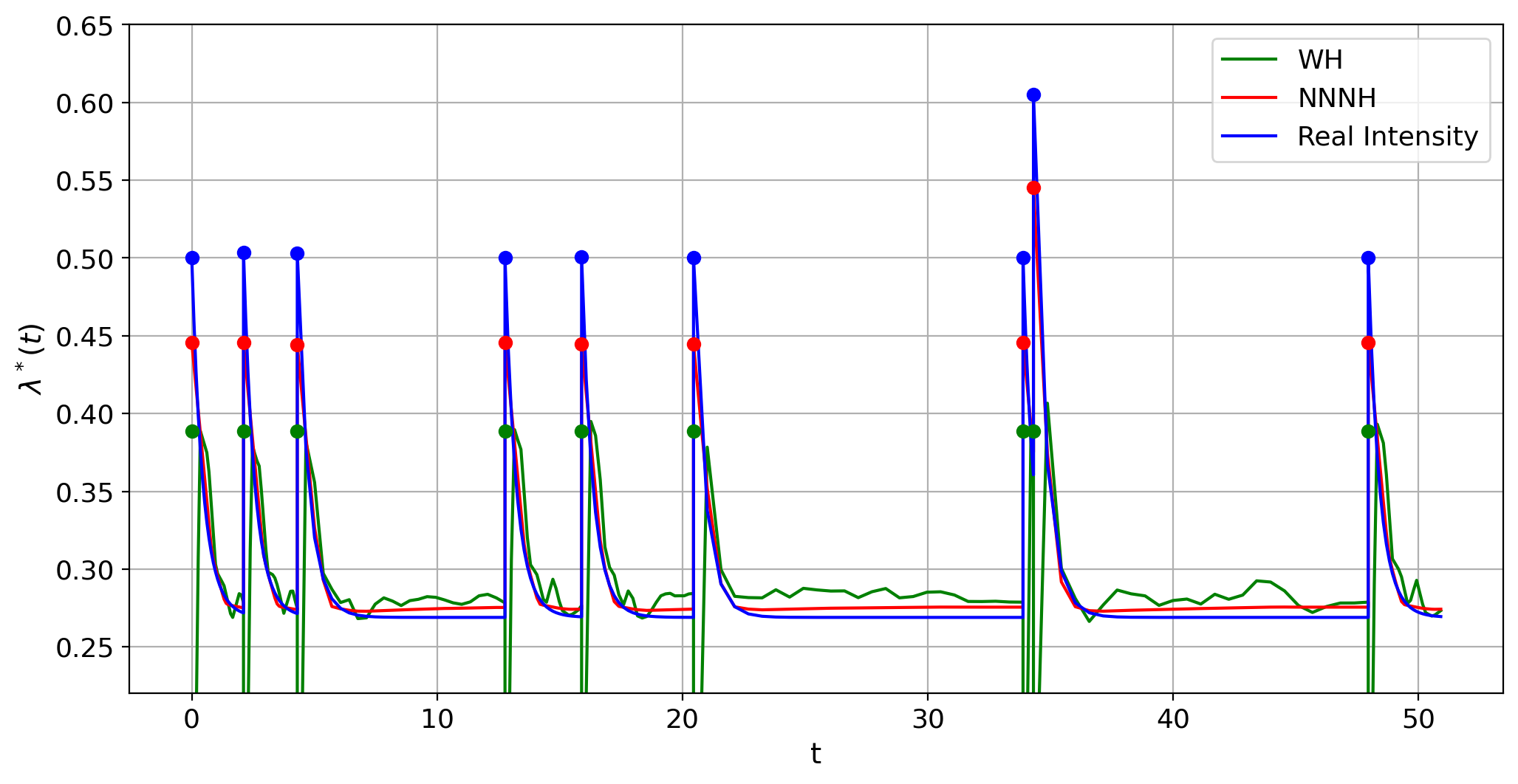}
		\caption{Comparison of real and estimated intensity function by WH and  NNNH for nonlinear Hawkes process with sigmoid as $\psi$ function.}
		\label{Fig:1dSigConsMu}
	\end{figure}

	The nonlinear Hawkes process was simulated for a duration of $T=10000$, resulting in $N(T)=3028$ events.  According to Equation \ref{Eq:multidim_HawkesApprox}, the NNNH model converts a non-linear Hawkes process with any $\Psi$ into a non-linear Hawkes process with $\Psi$ as a $\max$ function. As a result, the recovered kernels from the NNNH method might not correspond to the kernels of the original process. However, we can compare the actual simulated intensity process with the intensities recovered from the NNNH method and the WH method. Figure \ref{Fig:1dSigConsMu} demonstrates that the recovered intensities match the simulated intensities.
	
\begin{figure}[h!]
		\centering
		\includegraphics[width= 0.6\textwidth, height =0.3\textheight]{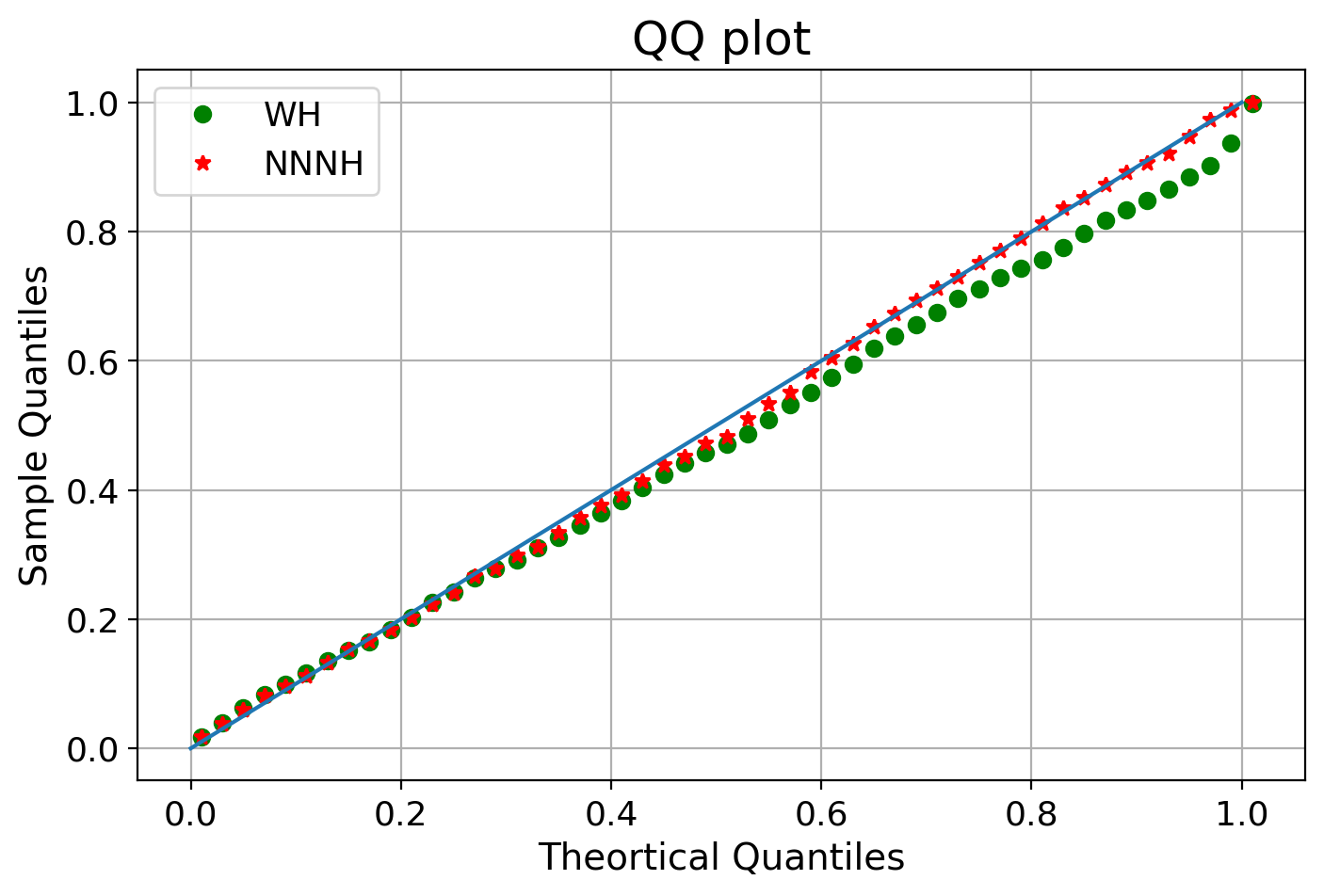}
		\caption{QQ plot comparing estimated and true intensities for sigmoid nonlinear Hawkes process using WH and NNNH}
		\label{Fig:QQplotSig}
	\end{figure}

	The accuracy of NNNH estimates of quantiles is evident from the QQ plot obtained from the fitted WH and the NNNH method on the test dataset, as shown in Figure \ref{Fig:QQplotSig}. The negative log-likelihood values for NNNH and WH are $2954,$ and $3813$ respectively.
	
	\paragraph{Estimation of Hawkes processes with varying base intensity function}

\begin{figure}[H]
		\centering
		\subcaptionbox{Base Intensity}{\includegraphics[width=0.50\textwidth,height=0.25\textheight]{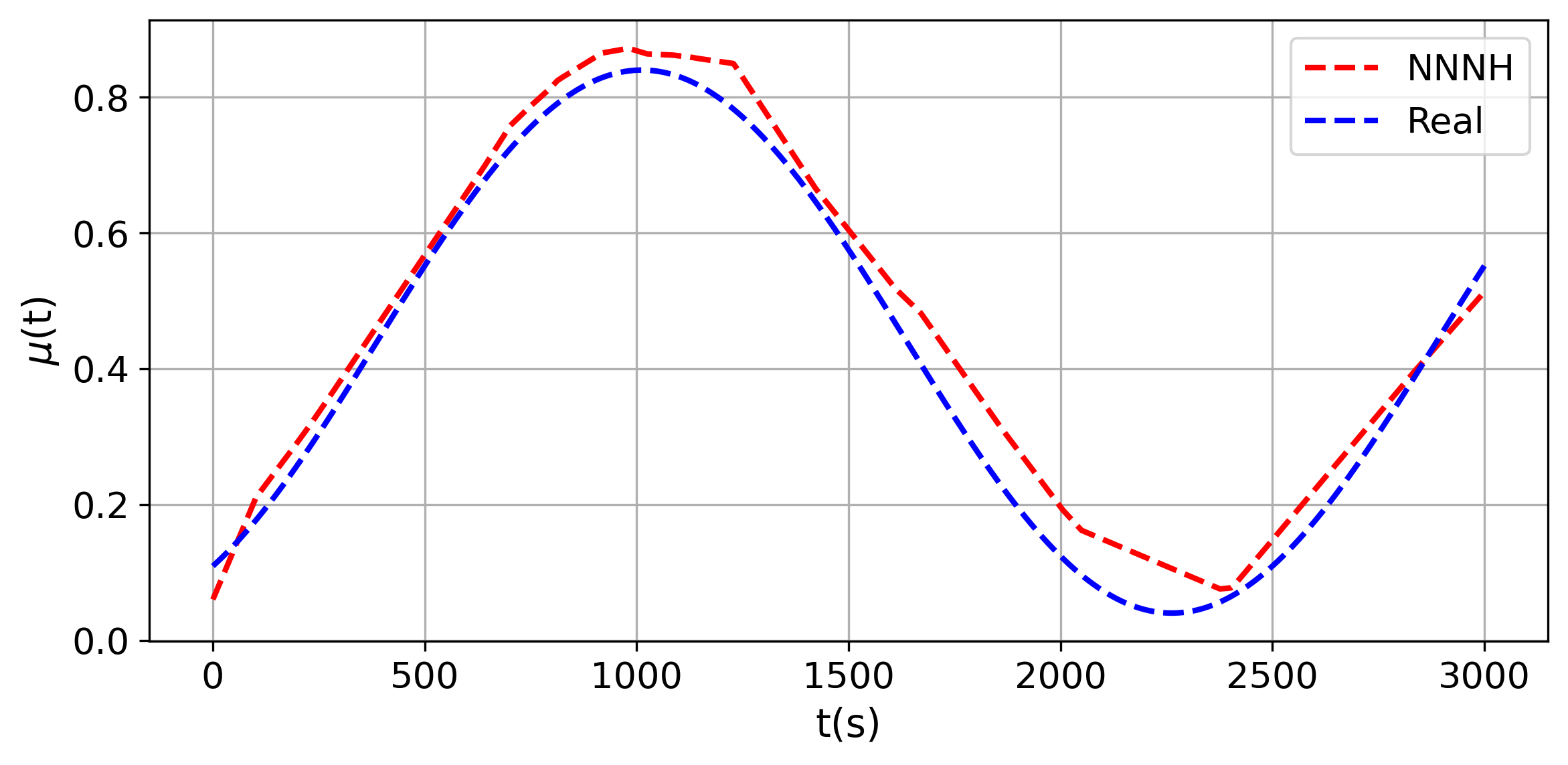}}%
		\subcaptionbox{Kernel}{\includegraphics[width=0.50\textwidth,height=0.25\textheight]{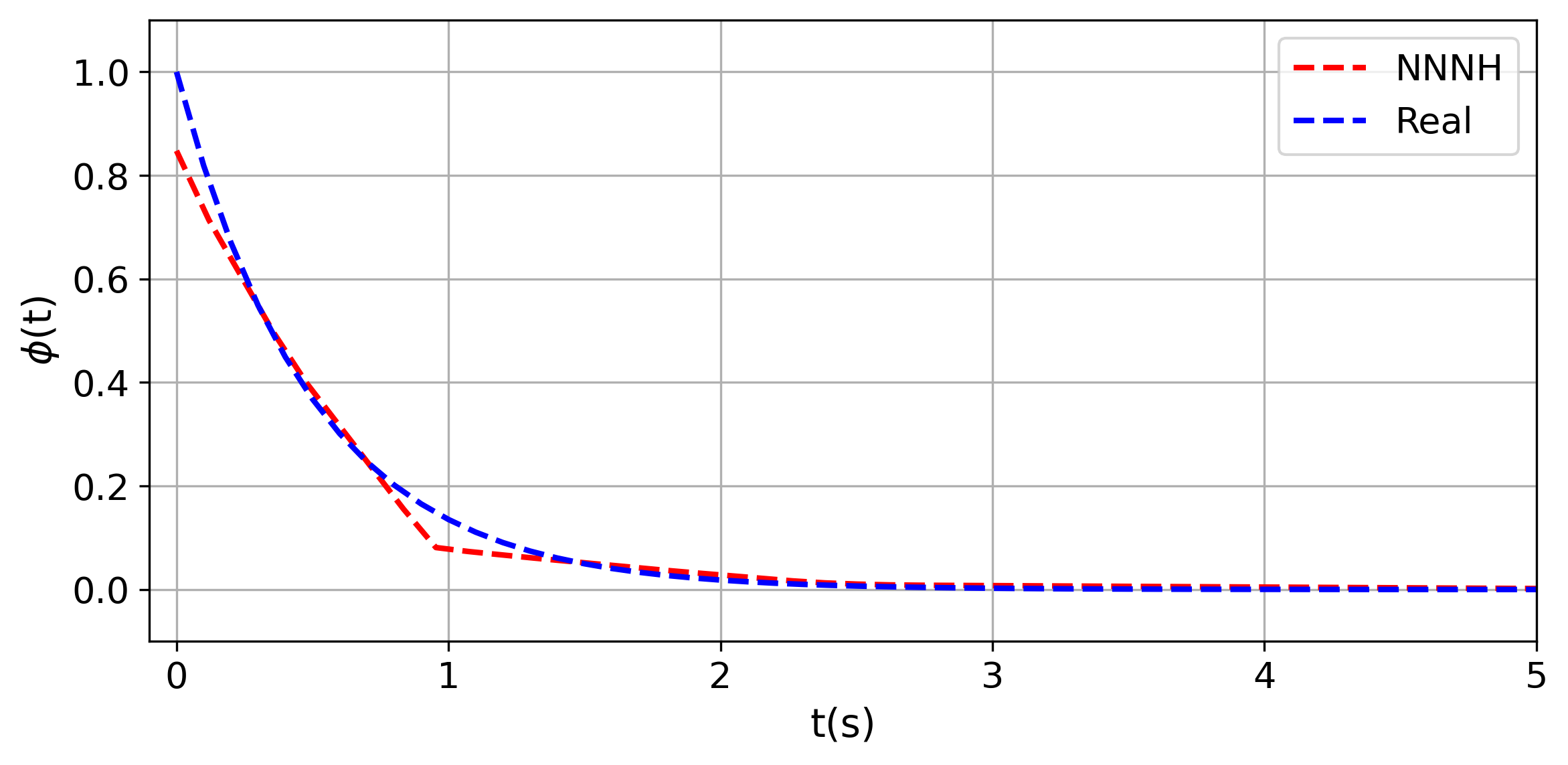}}%
		\hfill 
		\caption{ Estimated kernel and base intensity by applying NNNH to a one-dimensional Hawkes process with sin base intensity and an exponential kernel }
		\label{Fig:1dsinMu}
	\end{figure}

	To complete the univariate case test for the NNNH, we examine a Hawkes process with a base intensity that varies over time. We adopt a trigonometric sin function for the time-varying base intensity $\mu(t)$, which is specified in Table \ref{Tab:PPNLL}. The associated kernel for the Hawkes process is expressed as:
	$$
	\phi(t) = e^{-2 t }.$$
	We simulate the process till $T=3000,$ which results in $N(T)=2482$ events.  The comparison between the recovered base intensity function and the excitation kernel using the NNNH method with the true values is depicted in Figure \ref{Fig:1dsinMu}. The results demonstrate that the NNNH approach can accurately estimate both the base intensity function and the kernel concurrently.
	
	\subsection{Multivariate Hawkes Estimation}
	\label{Subsec:MultiHawk}
	
	In this section, we evaluate the NNNH approach for estimating multivariate non-linear Hawkes processes using a simulated dataset. The multivariate non-linear Hawkes process is simulated using the \textit{Ogata's thinning algorithm}. The parameters of the neural networks utilized to model the kernels in the NNNH method are optimized by maximizing the log-likelihood through the stochastic gradient descent method, as described in Section \ref{Sec:gradient}.  We utilize the WH method \citep{bacry2014second} and the Bonnet Multivariate method \citep{bonnet2022inference} as our comparative models. While the WH method is non-parametric, the Bonnet Multivariate approach assumes a parametric structure for the kernels in the Hawkes process. Both reference models are capable of estimating kernels that exhibit inhibitory effects.
	
	We consider two examples for the simulated case. We first consider a multivariate Hawkes process, with both positive and negative exponential kernels. Specifically the parameter chosen for the model are: 
	$$\alpha =  \begin{bmatrix}
		-0.9 & 3 \\
		1.2  & 1.5 
	\end{bmatrix},$$  $$\beta= \begin{bmatrix}
		5 & 5 \\
		8 & 8 
	\end{bmatrix}$$ , and a constant base intensity $$\mu =\begin{bmatrix}
		0.5\\
		1.0 
	\end{bmatrix}.$$ 
	
	The kernels of the process are defined as: $\phi_{dj}(t) = \alpha_{dj} e^{-\beta_{dj} t},\, 1\leq d,j \leq 2.$
	
\begin{figure}[H]
		\centering
		
		\includegraphics[clip,width=\columnwidth]{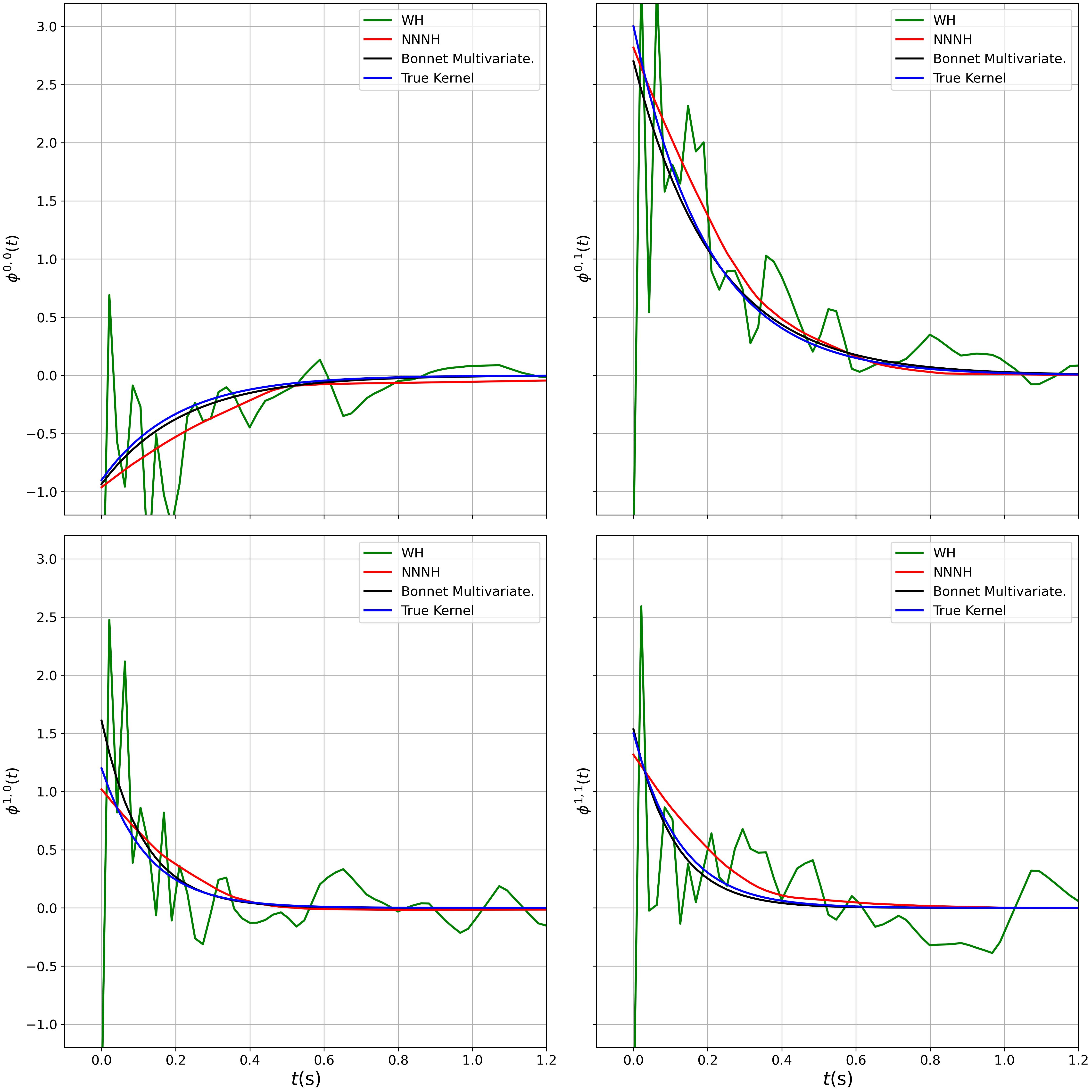}%
		
		\caption{The kernels fitted using the NNNH, WH, and Bonnet Multivariate methods, for a bivariate Hawkes process with both positive and negative kernels.}
		\label{Fig:2dNegKer}
	\end{figure}

	We simulate the process till $T=1000,$ which results in $N(T)=1002$ events. Figure \ref{Fig:2dNegKer} plots the true kernels and the kernels recovered using the NNNH, the WH, and the Bonnet Multivariate methods. Through a visual examination, it can be observed that the Bonnet Multivariate approach closely approximates the actual kernels. This outcome is in line with expectations as the simulated data employed a parametric form of the Hawkes process that is specifically suitable for the Bonnet Multivariate model. Regarding the non-parametric models, the WH estimates exhibit greater variability compared to the NNNH estimates. The respective negative log-likelihoods for the NNNH, the WH, and the Bonnet Multivariate methods are $1480$, $1967$, and $1460.$
	
	Subsequently, we examine a bivariate non-linear Hawkes process that encompasses a combination of diverse types of kernels. Precisely, we select the kernels as follows:

	$$ \textrm{exponential kernel}\,\phi_{11}(t) = 0.3 e^{-3t}, $$ 
	$$\textrm{rectangular kernel}	\,\phi_{12}(t) = 
	\left\{
	\begin{array}{l}
		0.7\times 0.4, \textrm{if } 1 \le t \le 1+\frac{1}{0.4}, \\
		0, \textrm{otherwise}
	\end{array},
	\right.$$
	$$\textrm{negative exponential} \, \phi_{21}(t)=-0.2e^{-t}, $$
	$$\textrm{exponential}\, \phi_{22}(t)=0.4e^{-2t}, $$
	
	and use the following base intensity values: 
	$$\mu =  \begin{bmatrix}
		0.1 \\
		0.2
	\end{bmatrix}$$ 
	
\begin{figure}[H]
		\centering
		
		\includegraphics[clip,scale=1,width=0.8\columnwidth]{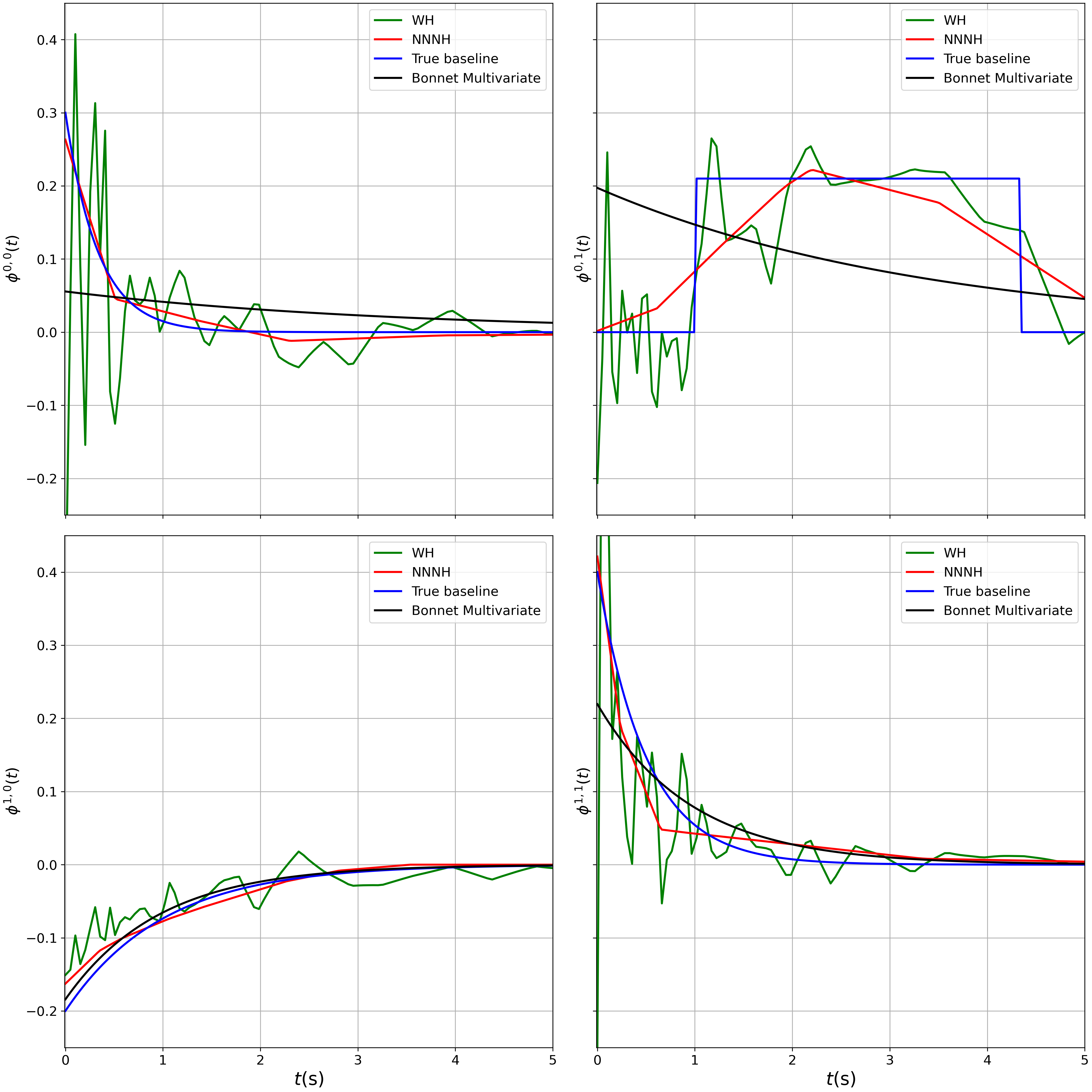}%
		
		\caption{ The fitted kernels, using the NNNH, WH, and Bonnet Multivariate method, for a bivariate non-linear Hawkes process with a mix of exponential and rectangular kernels. }
		\label{Fig:2dRectKernel}
		
	\end{figure}

	We perform simulations up to time $T=10000$, resulting in the occurrence of a total $8489$ events in the two dimensions. Figure \ref{Fig:2dRectKernel} presents the findings obtained from the NNNH approach in comparison with other techniques. Notably, the advantages of employing a non-parametric estimation method for Hawkes processes are evident in this context. Both the NNNH and the WH methods successfully capture all the kernels, whereas the Bonnet Multivariate approach does not accurately depict the rectangular kernel. It is worth mentioning that the estimated kernels obtained using the WH method appear to be more erratic than those of the NNNH. The corresponding negative log-likelihood values for the WH, the Bonnet Multivariate, and the NNNH are $3051,$ $3062,$ and $2899$ respectively.
	
	\subsection{Sensitivity Analysis}\label{Sec:sens}
	
	In the context of the NNNH approach, we have made deliberate decisions concerning the quantity of neurons deployed in the network's hidden layer to emulate the kernels and the learning rate employed to regulate the parameter updates in the Adam algorithm. To gauge the sensitivity of the NNNH method, we have conducted a numerical investigation with regard to these decisions. Specifically, we have examined the influence of variations in the number of neurons and learning rates on the negative log-likelihood value, while holding all other factors constant. Note that the likelihood values reported are for the validation dataset which was not used in the training of the network 
	
	The sensitivity analysis is conducted for the negative exponential kernel, which is discussed in Section \ref{Subsec:UniEst}. The study focuses on the variation in the number of neurons employed in the neural networks, and the corresponding impact on the negative log-likelihood values. The results are presented in Table \ref{Tab:hyperParamsNLL}. The findings indicate that alterations in the number of neurons have an insignificant effect on the minimum negative log-likelihood values obtained, but they do impact the computational time required to achieve optimal parameters. The neural network with thirty two neurons has the lowest computational time. Therefore, we use a neural network with 32 neurons in our experiments to model the kernels.

	\begin{table}[!h]
		\centering
		\caption{Percentage change in negative log-likelihood (NLL) values and corresponding computational time for different numbers of neurons in the neural network}
		\begin{tabular}{ |c|c|c|c| }
			\hline
			No of Neurons & NLL & $\%$ change & $\%$ time change\\ 
			\hline
			2 & 2057.78 & -0.061 & +164.1\\
			4 & 2058.01 & -0.057 & +95.4\\ 
			8 & 2059.06 & -0.004  & +89.9\\ 
			16 & 2059.14 & -0.001 &+90.9 \\
			32 & 2059.16 & 0.00 & 0.00\\ 
			64 & 2061.29 & +0.10  & +6.7\\
			128 & 2061.63 & +0.12& +77.3 \\
			256 & 2061.62 & +0.12 & +76.1 \\ 
			\hline
		\end{tabular}
		\label{Tab:hyperParamsNLL}
	\end{table}
	Table \ref{Tab:hyperParamsLR} presents the percentage variation in the negative log-likelihood values linked to the optimal parameters obtained by varying the learning rate for the output layer of the neural network. The learning rate for the hidden layer remains unchanged, and is taken as ten percent of the learning rate for the output layer.
	
	\begin{table}[!h]
		\centering
		\caption{Percentage change in negative log-likelihood (NLL) values for different learning rates.}
		\begin{tabular}{ |c|c|c| }
			\hline
			Learning Rate & NLL & $\%$ change \\ 
			\hline
			0.01 & 2064.54 & +0.00 \\
			0.005 &  2064.19  & -0.02 \\ 
			0.001 & 2065.49 & +0.05 \\
			0.0005 & 2067.87 & +0.12 \\ 
			\hline
		\end{tabular}
		\label{Tab:hyperParamsLR}
	\end{table}
	
	We find that the optimal network parameters obtained using varying learning rates results in similar negative log-likelihood values and therefore use a learning rate of 0.01 for all our experiments.

	\subsection{Real Data}\label{Subsec:realdata}

	\subsubsection{Financial Dataset} \label{Subsubsec:findata}
	
	In this study, we evaluate the efficacy of the NNNH method on high-frequency order book data pertaining to two of the most frequently traded cryptocurrencies, namely bitcoin and Ethereum. The data comprises of buy and sell trade records for the BTC-USD (bitcoin- US dollar) and ETH-USD (Ethereum- US Dollar) pairs. We streamed the Binance exchange order book data, as several popular cryptocurrencies are traded in this exchange, and the exchange has high trade volumes. We obtain the tick-by-tick arrival times for both buy and sell trades from the exchange. The market is composed of makers and takers, with makers generating buy or sell orders that are not carried out immediately, thereby creating liquidity for that cryptocurrency. In contrast, takers place market orders that are executed instantly, thereby taking away the liquidity. 
	
	The Binance exchange furnishes two streams of tick-by-tick data: trade arrival data and trade stream data. The trade arrival data comprises of limit orders, which are orders placed with a specified price limit, and market orders, which are orders executed at the prevailing market price. Limit orders are executed when the best available market price reaches the set limit, while market orders are executed instantly at the current best limit order. The Binance trade-stream data provides the timestamp of these order arrivals, along with price and volume features, and a unique identifier for the buyer/seller. Since a single market order may necessitate multiple limit orders to fulfill the requested volume, several trades are recorded with a common identifier. Therefore, we cleaned the dataset by filtering out the data with common IDs and retaining only the unique trade events. Finally, based on whether the buyer was a market maker or taker, the trades were marked as either a buy or a sell market order.
	
	Our analysis focuses exclusively on the trades conducted for the BTC-USD and ETH-USD pairs, which account for the predominant trading volumes in the cryptocurrency exchange. The arrival time for market orders for these two pairs can be modelled as a four dimensional non-linear Hawkes process, i.e.,
	\begin{itemize}
		\setlength{\itemindent}{5em}
		\item[First Dimension:] Intensity process for the sell market orders for the BTC-USD pair
		\item[Second Dimension:] Intensity process for the buy market orders for the BTC-USD pair
		\item[Third Dimension:] Intensity process for the sell market orders for the ETH-USD pair
		\item[Fourth Dimension:] Intensity process for the buy market orders for the ETH-USD pair.
	\end{itemize}
	
	The analysis was performed for the market orders arrival times on December 7 2021, between 12:00  to 12:10 (UTC) on the Binance exchange. Table \ref{BTCSum} summarises the data after classifying the trades as buy or sell market order events.

	\begin{table}[!h]
		\centering
		\caption{Summary of the crypto-currency market orders recorded on December 7 2021 between 12:00 and 12:10 UTC}
		\begin{tabular}{ |c|c|c|c| }
			\hline
			Market Order Type&BTC-USD & ETH-USD & Aggregate \\ 
			\hline
			Sell &4219&3480& 7699\\
			Buy &4374&2999& 7373\\\hline
			Total &8593&6479&15072\\
			\hline
		\end{tabular}
		\label{BTCSum}
	\end{table}
	
\begin{figure}[h!]
		\centering									
		
		\includegraphics[height=0.4\textheight,width=\textwidth]{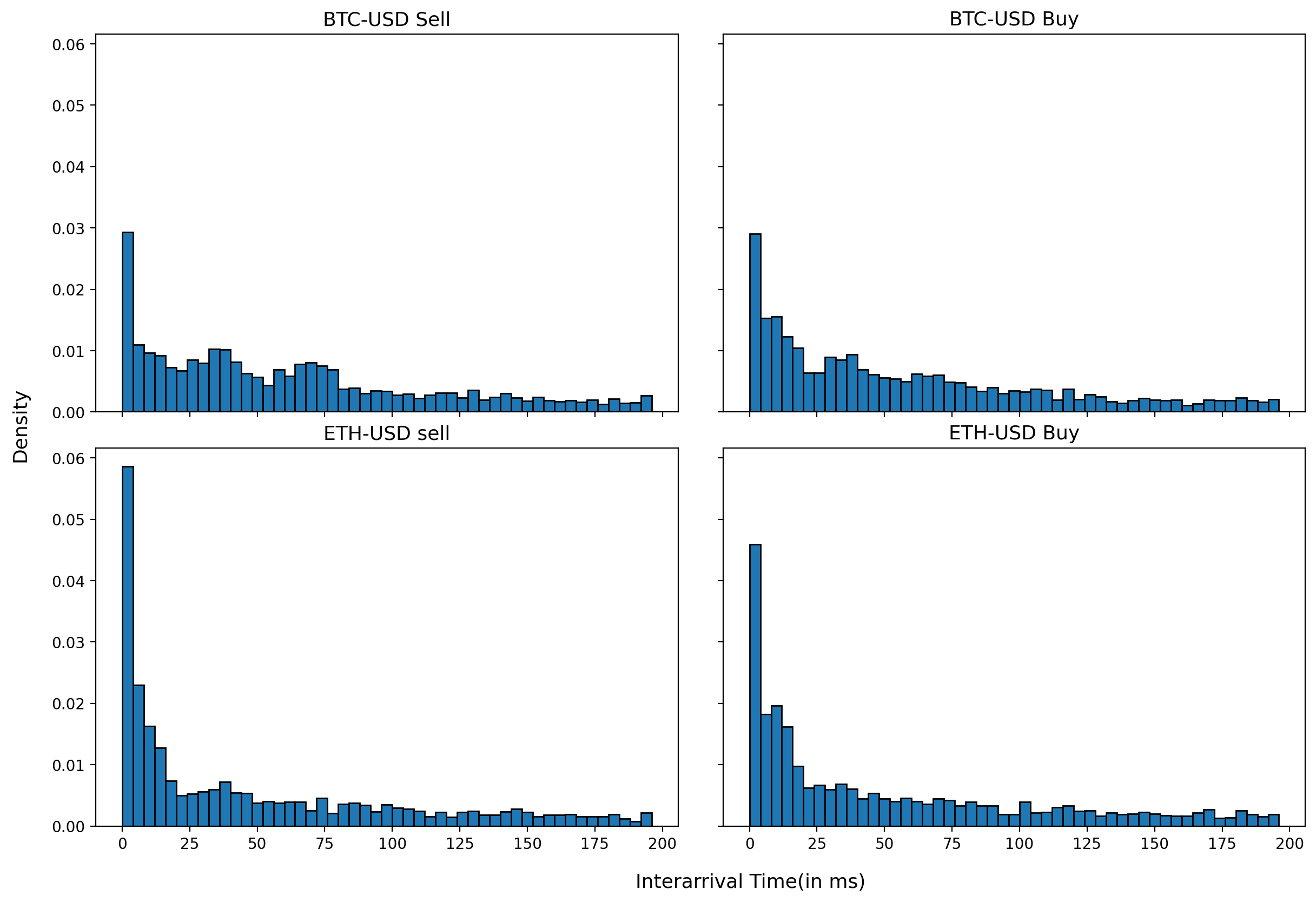}
		
		\caption{Histogram for the inter event time arrivals of the BTC-USD sell trade}
		\label{Fig:btceth_hist}
	\end{figure}

	Figure \ref{Fig:btceth_hist} provides the histogram of the inter-arrival time of the sell BTC-USD market orders. The timestamps for order arrival are in milliseconds.
	
	Our aim is to employ the NNNH method to jointly model the intensity process of buy and sell market orders for BTC-USD and ETH-USD currency pairs. This modelling approach enables us to gain insights into the causal relationships between the two pairs, such as whether the arrival of a buy BTC-USD order affects the arrival of a sell ETH-USD market order. Furthermore, identifying the functional form of the self and cross modulation due to the arrival of market orders is of interest. This highlights the necessity of non-parametric estimation techniques, as the true form of the modulation function remains unknown. To achieve our objective, we partition the dataset into train, validation, and test sets. We optimize the parameters of the model using stochastic gradients computed on the train set and use the validation set for applying the early stopping criteria. Specifically, if there is no improvement in negative log-likelihood values computed on the validation set for ten consecutive iterations, we stop training. Finally, we evaluate the goodness of fit using the negative log-likelihood reported on the test set.
	
\begin{figure}[H]	
		\centering
		\includegraphics[height=0.5\textheight,width=\textwidth]{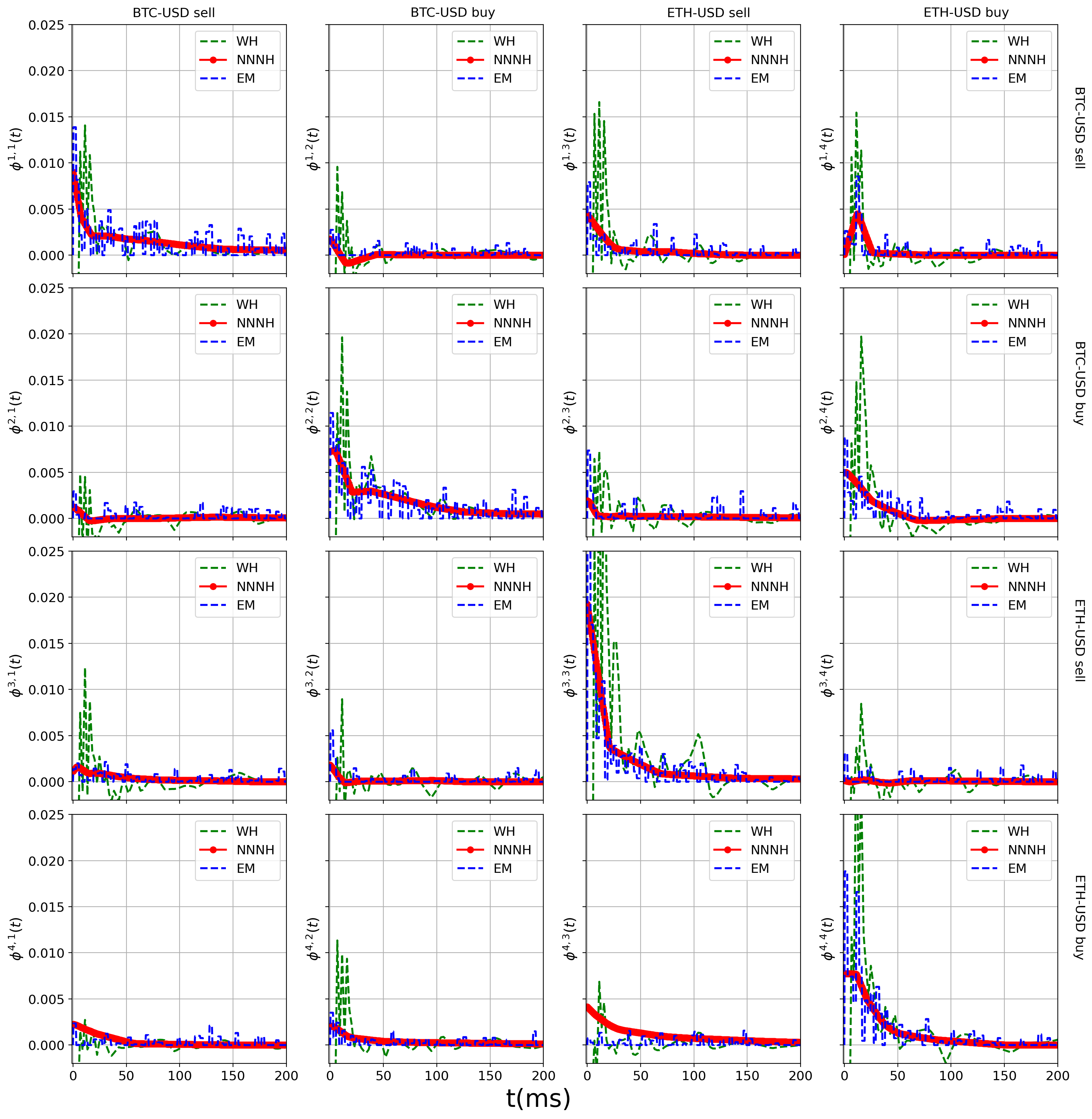} 	
		
		\caption{Fitted kernels for BTC-USD and ETH-USD sell-buy market orders estimated using the NNNH, the WH, and the EM methods.}
		\label{Fig:btceth_KerAll}
	\end{figure}

	We adopt the widely used non-parametric methods, namely the WH and the EM method \citep{lewis2011nonparametric}, as reference models . In Figure \ref{Fig:btceth_KerAll}, we compare the obtained kernels using the three methods. The corresponding negative log-likelihood values are $9630$, $6732$, and $6400$ for the EM, WH, and NNNH methods, respectively. The estimated base intensities for BTC-USD sell, BTC-USD buy, ETH-USD sell, and ETH-USD buy, in the order of appearance, using the NNNH method are $0.0028$, $0.0032$, $0.0021$, and $0.0022$, respectively.
	
	By examining Figure \ref{Fig:btceth_KerAll}, we can draw the following observations for the market orders that arrived during the training window:
	
	\begin{itemize}
		\item Significant self-excitation is observed for both BTC-USD and ETH-USD buy and sell market orders.
		\item BTC-USD sell orders result in excitation of ETH-USD buy and sell orders.
		\item The self-excitation of ETH-USD buy and sell orders is higher compared to their respective BTC-USD orders.
		\item A certain level of inhibition is observed in BTC-USD buy orders due to sell BTC-USD orders.
	\end{itemize}
	
	Notably, all three methods effectively capture a lagged cross-excitation effect in ETH-USD buy orders caused by sell BTC-USD orders.

	\paragraph{Predictive capacity of the NNNH method}
	
	Accurately inferring the kernels is critical in developing an improved predictive model for the arrival process. For example, a trading algorithm relies on predicting the time at which the next market order will arrive with 90\% confidence. With an accurate prediction model, approximately 90\% of the predictions are expected to be correct, and in roughly 10\% of the cases, the order will arrive after the predicted time. If the model consistently produces accurate predictions (i.e., does not fail for 10\% of the cases), then the predictions are overly conservative, indicating that the predicted time is set too far in the future. Conversely, if more than 10\% of the predictions fail, the predicted time is closer than anticipated. Therefore, a prediction model that produces predictions with Q\% certainty should have Q\% correct outcomes. The accuracy of such a prediction model can be evaluated using a QQ plot.

\begin{figure}[h!]
		\includegraphics[height=0.5\textheight,width=\textwidth]{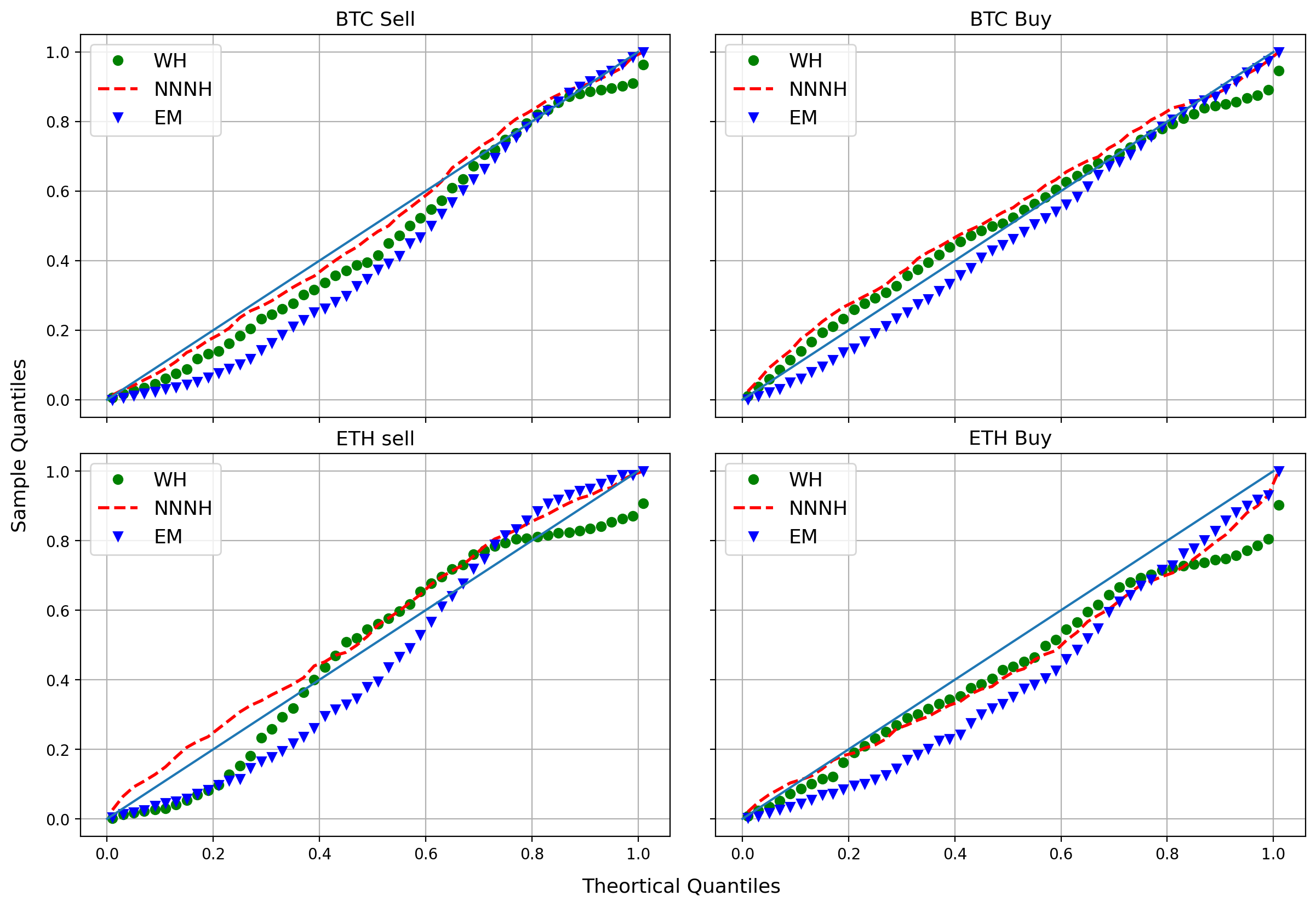}
		\caption{QQ plot comparing order arrivals of BTC-USD and ETH-USD using EM, WH, and NNNH methods}
		\label{Fig:QQplot}
	\end{figure}

	Figure \ref{Fig:QQplot} presents the QQ plot for the BTC-USD and ETH-USD order arrivals based on the EM, WH, and NNNH models. The sample quantiles for the QQ plot are obtained from the test dataset. The results indicate that all three models make reasonably accurate predictions. Notably, the NNNH method appears to have more precise predictions, particularly for BTC-USD and ETH-USD sell orders, as evidenced by the visual assessment.
	
	\subsubsection{Neuron Spike Train Dataset}\label{Subsubsec:spikedata}
	
	In this study, we utilized the NNNH method to analyze a dataset of neuron spikes, which was obtained from an experiment conducted by \citet{engelhard2013inducing}. The primary objective of the experiment was to examine the relationship between unit recordings from the motor cortex of monkeys and the position and grip strength of their hands as they utilized a joystick to manipulate a robotic arm. Consistent with the approach taken by \citet{aljadeff2016analysis}, we analyzed the data obtained from the nine simultaneous electrode recordings. Although the dataset included additional information regarding hand motion, such as cursor trajectory and grip force, our analysis was focused exclusively on the neuron spike train.
	
	To provide a brief biological background about neuron spikes, we refer to \cite{DUVAL2022180}. Neurons employ an electrical wave of short duration, known as the action potential or spike, to reliably transmit signals from one end to the other \citep{castelfranco2016evolution}. Action potentials are all-or-none events and are triggered when the neuron's membrane potential, which is the electrical potential difference between the inside and the outside of the neuron, is sufficiently large. Between two successive action potentials, the neuron adds up its inputs, causing changes in its membrane potential. When this potential reaches a sharp threshold, the neuron fires a spike that propagates \citep{luo2015inferring}.

\begin{figure}[H]
		\centering
		\includegraphics[width=0.7\textwidth,height=0.3\textheight]{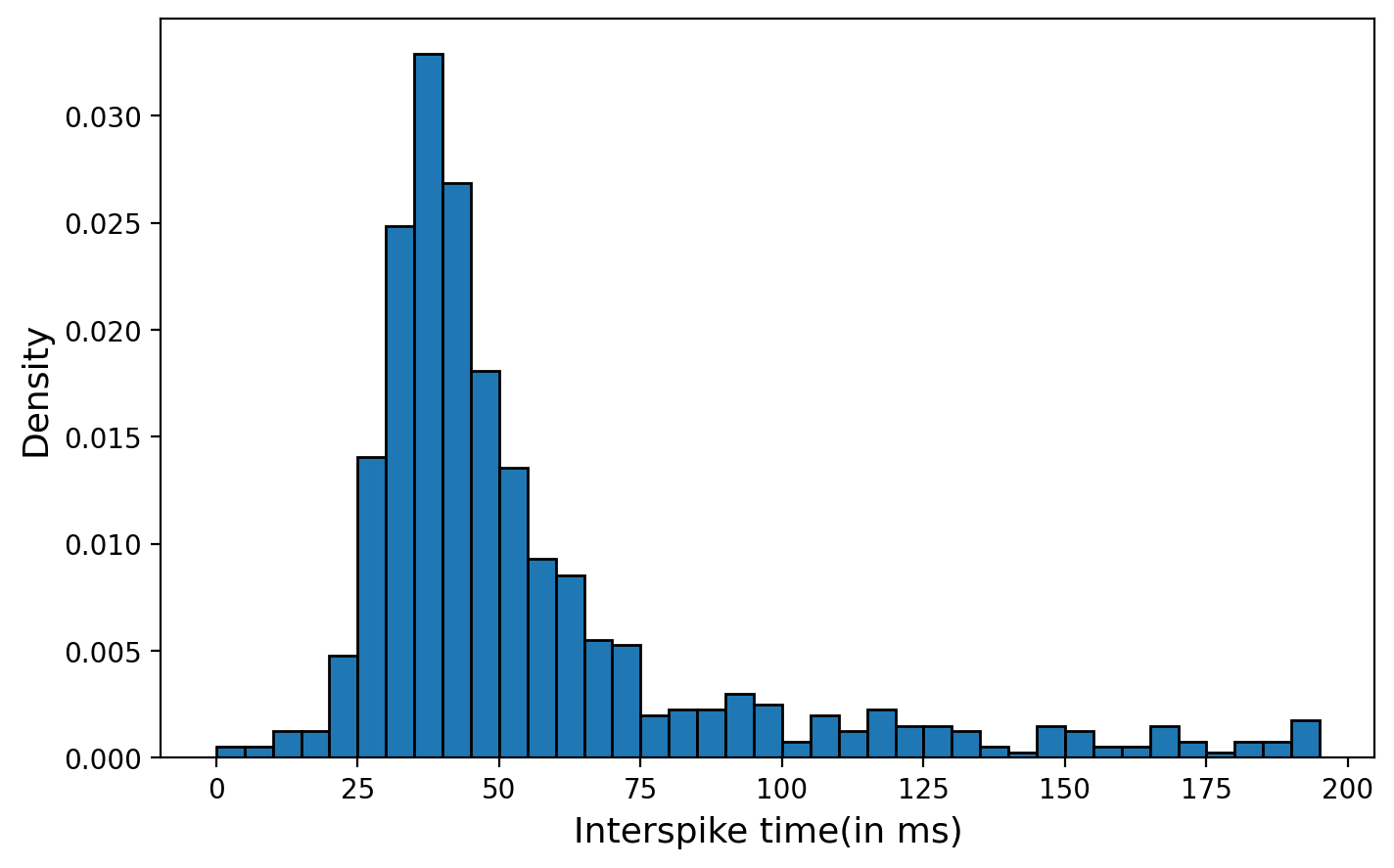}
		\caption{Histogram of inter-spike time intervals for neuron 1 in grip-and-reach task}
		\label{Fig:neuronhist}
	\end{figure}

	The histogram in Figure \ref{Fig:neuronhist} presents the inter-arrival times of spikes generated by a single neuron. Notably, the peak of the histogram is marginally displaced from the origin, in contrast to the histogram of a conventional exponential distribution. This observation suggests the presence of an inhibitory process. To model the neuron spike train, we employed a non-linear Hawkes process that is nine-dimensional. Each dimension of the process is associated with the intensity of neuron spike arrivals at an individual electrode. We use the NNNH method to estimate the kernel functions and the base intensities for this process. 
	
 Table\ref{Tab:neuronCounts} gives the  number of spike events for each neurons used for the analysis. 
	\begin{table}[H] 
		\centering
		\caption{Summary statistics of neuron spike data}
		\begin{tabular}{ |c|c|c|c|c||c|c|c|c|c| } 
			\hline
			Neuron&$1$ & $2$ & $3$ & $4$ & $5$ & $6$ & $7$ & $8$ & $9$\\
			\hline \hline
			No. of events & 4121& 1895 & 912 & 2509 & 221 & 2784 & 149 & 3854 & 1360\\
			\hline 
		\end{tabular}
		
		\label{Tab:neuronCounts}
	\end{table}
	
\begin{figure}
		\includegraphics[height=\textwidth,width=\textwidth]{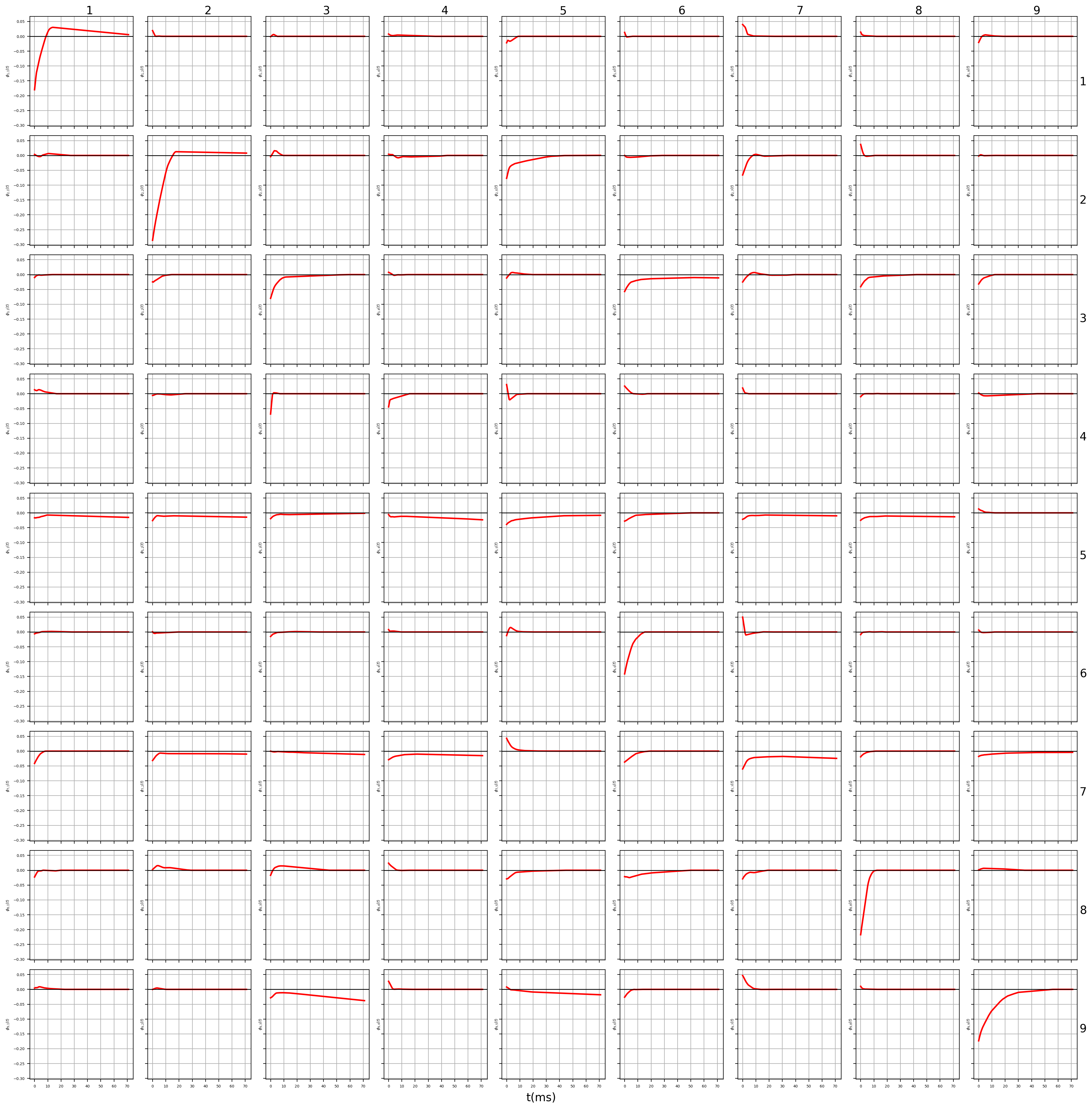}
		\caption{Kernels of motor neurons demonstrating the self-dependency and mutual dependency of neurons for the grip-and-reach task estimated by NNNH.}
		\label{Fig:neuronKer}
	\end{figure}

	Figure \ref{Fig:neuronKer} displays the kernels that were fitted using the NNNH method to model the neuron spike train. Additionally, Table \ref{Tab:neuronBase} reports the base intensity values for the arrival of neuron spikes at each electrode. The results indicate that a majority of the neurons exhibit a self-inhibitory characteristic. Specifically, after firing, neurons typically undergo a refractory period during which they cannot generate additional spikes. The inhibitory pattern evident in the plots likely corresponds to this refractory period. Based on the figures, it is reasonable to assume that the refractory period for the recorded motor neurons falls within the 20-60 ms range.

	\begin{table}[H] 
		\centering
		\caption{Base intensities of neuron spike data obtained using the NNNH for the grip-and-reach task}
		\begin{tabular}{ |c|c|c|c||c|c|c|c|c| } 
			\hline
			$\mu_1$ & $\mu_2$ & $\mu_3$ & $\mu_4$ & $\mu_5$ & $\mu_6$ & $\mu_7$ & $\mu_8$ & $\mu_9$\\
			\hline \hline
			0.009& 0.035&0.001&0.046&0.001& 0.043&0.001&0.076&0.022\\
			\hline 
		\end{tabular}
		
		\label{Tab:neuronBase}
	\end{table}
	
	\section{Conclusion}\label{Sec:conclusion}
	
	This paper proposes a non-parametric method for estimating non-linear Hakes processes. We call the method  Neural Network for Non-linear Hawkes  processes or the NNNH method. The NNNH method involves modelling the kernels and the base intensity functions of a non-linear Hawkes process as individual neural networks. The parameters of the neural networks involved are jointly estimated by maximizing the log-likelihood function using the batch stochastic gradient descent with Adam for adaptive learning. To apply the SGD method, an unbiased estimator for the gradients with respect to the network parameters is proposed. The paper provides an efficient scheme to evaluate the integrated intensity and its gradients, which can be a computational bottleneck.
	
	The effectiveness of the NNNH method in learning the underlying process is investigated through numerical experiments. Unlike many recent neural network-based models that predict conditional intensities, the NNNH model retains the interpretability of the Hawkes process by enabling the inference of the kernels rather than just direct prediction of the conditional intensities. The ability to recover the kernels is desirable for understanding causal relationships between the arrivals in different dimensions, for instance. The NNNH method performed satisfactorily for the diverse set of problems considered. We also see that the NNNH can be used to estimate non-homogeneous Poisson process.  The NNNH method is able to recover non-smooth kernels, kernels with negative intensities, and estimate other variants of the non-linear Hawkes processes. 
	
	The NNNH approach is applied to examine the process of order arrivals for buy and sell transactions on the Binance exchange in relation to the BTC-USD and ETH-USD currency pairs. We find evidence of both self and cross excitation in the order arrivals for the two currency pairs. Notably, self excitation is observed to be significant for both currency pairs, while the cross dependence exhibits an asymmetry whereby ETH-USD order arrivals are more strongly influenced by BTC-USD order arrivals than vice versa. 
	
	We applied the NNNH model to a publicly accessible dataset of neuron spikes obtained from the motor cortex of monkeys engaged in a grip-and-reach task. As anticipated, the kernels obtained from the analysis exhibit self-inhibition, which could be associated with the refractory period of the neurons. Our findings provide evidence of the NNNH method's efficacy in analyzing high-dimensional datasets.
	
	The NNNH method has a disadvantage in that the estimation of model parameters relies on the use of stochastic gradient descent (SGD), which updates the network parameters iteratively using first-order derivatives. Convergence of SGD-based methods can be slow, often requiring several iterations before the stopping criterion is met. However, the use of SGD makes the NNNH method well-suited for online learning, enabling the model to be trained on new data points that are continuously streaming in, as opposed to a static dataset. Moreover, the NNNH method is amenable to parallel computing of the gradients for batch SGD. A potential avenue for future research is to extend the NNNH method to model marked non-linear Hawkes processes.

	\medskip
	
	\small
	\bibliographystyle{plainnat}
	\bibliography{notes_1}
	\normalsize
	\newpage 
	\appendix
	\section*{Appendix}
	\section{Gradients of the Neural Network Parameters}\label{App:gradKernel}
	As both $\mu_d(x)$ and $\phi_{dj}(x)$ has similar network, the gradients will also be similar. The gradient with respect to each of the parameter$\theta_p$ is given by,
	\begin{equation*}
		\nabla_{\theta_p}\widehat{\phi}_{dj}(x) =  \nabla_{[b_2,a^p_2,a^p_1,b^p_1]} \left(b_2+ \sum_{i=1}^{P}a^i_2  \max\left(a^i_1 x+b^i_1,0\right)\right)
	\end{equation*}
	The gradients corresponding to each of the neural parameters were provided as,
	\begin{align*}
		\nabla_{b_2} {\phi}_{dj}(x) &= 1 \\
		\nabla_{a_2^p} {\phi}_{dj}(x) &= \max\left(a^p_1 x+b^p_1,0\right) \\
		\nabla_{a_1^p} {\phi}_{dj}(x) &= \left(a_2^px\right)\mathbbm{1}_{\left(a^p_1 x+b^p_1,0\right)>0} \\
		\nabla_{b_1^p} {\phi}_{dj}(x) &= \left(a_2^p\right)\mathbbm{1}_{\left(a^p_1 x+b^p_1,0\right)>0}
	\end{align*}
	
	\section{Integrated Intensity Function - Network Representation}\label{App:integralExpression}
	Given the splitting criteria in section\ref{Subsec:IntLambda},
	$$
	\int_{s_{q-1}}^{s_q} \widehat{\lambda}^*_d(s) ds = \int_{s_{q-1}}^{s_q} \max\left(\widehat{\lambda}_d(s), 0\right) ds,
	$$
	If there are no zero crossings for $\max\left(\widehat{\lambda}_d(s), 0\right)$ within the interval $[s_{q-1},s_q]$, then $\widehat{\lambda}_d^*(s)$ is linear in the interval. Also if $\widehat{\lambda}_d(s_{q-1}) >0$  , then the network expression for the integrated intensity is given as,
	\begin{align*}
		\int_{s_{q-1}}^{s_q} \max\left(\widehat{\lambda}_d(s), 0\right) ds& = \left[\int_{s_{q-1}}^{s_q} \widehat{\lambda}_d(s) ds\right]\mathbbm{1}_{\widehat{\lambda}_d(s_{q-1}) >0} \\ 
		&= \left[\int_{s_{q-1}}^{s_q} \widehat{\mu} +\sum_{\{k,i,j\}} a^i_2  \max\left(a^i_1 (s-t_k^j)+b^i_1,0\right)\right] \mathbbm{1}_{\widehat{\lambda}_d(s_{q-1}) >0} \\
		&= \left[\int_{s_{q-1}}^{s_q} \widehat{\mu} +\sum_{\{k,i,j\}} a^i_2  \left[a^i_1 (s-t_k^j)+b^i_1\right] \mathbbm{1}_{a^i_1 (s-t_k^j)+b^i_1>0}\right] \mathbbm{1}_{\widehat{\lambda}_d(s_{q-1}) >0} \\
		&=\left[ \widehat{\mu}s +\sum_{\{k,i,j\}} a^i_2  \left[a^i_1\left(\frac{s^2}{2}-t_k^j s\right) 
		+ b^i_1 s\right] \mathbbm{1}_{a^i_1 (s-t_k^j)+b^i_1>0}\right]^{s_q}_{s_{q-1}} \mathbbm{1}_{\widehat{\lambda}_d(s_{q-1}) >0}
	\end{align*}

\end{document}